%% file: main.tex
\documentclass{article}

\PassOptionsToPackage{numbers, compress}{natbib}
\usepackage[preprint]{arxiv}
\usepackage[utf8]{inputenc}
\usepackage[T1]{fontenc} 
\usepackage{url}
\usepackage{booktabs}
\usepackage{amsfonts}
\usepackage{nicefrac}
\usepackage{microtype}
\usepackage{xcolor}
\usepackage{graphicx}
\usepackage{xspace}
\usepackage{amsmath}
\usepackage{amssymb}
\usepackage{multirow}
\usepackage{colortbl}
\usepackage{hhline} 
\usepackage{pifont}
\usepackage{soul}
\usepackage{dsfont}
\usepackage{tikz}
\usepackage{comment}
\usepackage{subcaption}
\usepackage{arydshln}
\usepackage{hyperref}
\input{def.tex}

\title{\texorpdfstring{SceneAligner: 3D-Grounded \\ Floorplan Localization in the Wild}{SceneAligner: 3D-Grounded Floorplan Localization in the Wild}}

\author{Junhyeong Cho$^{1}$ \quad Ruojin Cai$^{2}$ \quad Hadar Averbuch-Elor$^{1}$\\[2pt]
$^{1}$Cornell University \quad $^{2}$Kempner Institute, Harvard University\\[2pt]
{\small{\url{https://Cornell-VAILab.github.io/SceneAligner}}}}

\begin{document}

\maketitle

\input{sec/0_abstract}
\input{sec/1_introduction}
\input{sec/2_related_work}
\input{sec/3_method}
\input{sec/4_experiments}
\input{sec/5_conclusion}

\bibliography{main}
\bibliographystyle{plainnat}

\clearpage
\appendix 
\input{sec/supp_0_intro}
\input{sec/supp_1_dataset_statistics}
\input{sec/supp_2_implementation_details}
\input{sec/supp_3_qualitative_results}
\input{sec/supp_4_ablation_study}
\input{sec/supp_5_analysis}
\input{sec/supp_6_limitation}
\input{sec/supp_7_application}
\input{sec/supp_8_broader_impacts}

\end{document}

%% file: def.tex
\definecolor{darkblue}{rgb}{0, 0.2, 0.6}
\definecolor{darkgray}{rgb}{0.1, 0.1, 0.1}
\definecolor{gray}{rgb}{0.2, 0.2, 0.2}
\definecolor{lightgray}{rgb}{0.3, 0.3, 0.3}
\definecolor{deepdarkblue}{rgb}{0, 0.05, 0.3}
\definecolor{orange}{rgb}{1.0, 0.5, 0.0}
\definecolor{darkorange}{rgb}{0.7, 0.3, 0.0}
\definecolor{realred}{rgb}{1.0, 0.0, 0.0}
\definecolor{pastelred}{rgb}{0.9, 0.2, 0.45}
\definecolor{purple}{rgb}{0.6, 0.0, 0.6}
\definecolor{dogwoodrose}{rgb}{0.84, 0.09, 0.41}
\definecolor{mint}{rgb}{0.01176, 0.5490, 0.5490}
\definecolor{blue}{rgb}{0, 0, 1.0}
\definecolor{azure}{rgb}{0.0, 0.5, 1.0}
\definecolor{nicegreen}{rgb}{0.0, 0.7, 0.1}
\definecolor{CuGray}{gray}{0.9}
\definecolor{amethyst}{rgb}{0.6, 0.4, 0.8}
\definecolor{black}{rgb}{0.0, 0.0, 0.0}
\definecolor{steelblue}{rgb}{0.27, 0.51, 0.7}
\definecolor{brightcerulean}{rgb}{0.11, 0.67, 0.84}
\definecolor{brown}{rgb}{0.4, 0.3, 0.3}
\definecolor{mainblue}{rgb}{0.12,0.49,0.85}

\makeatletter
\DeclareRobustCommand\onedot{\futurelet\@let@token\@onedot}
\def\@onedot{\ifx\@let@token.\else.\null\fi\xspace}

\def\eg{\emph{e.g}\onedot}

\makeatother

\newcommand{\comparowseven}[1]{
    \includegraphics[width=0.133\linewidth]{figures/qualitative_comparison/Ours/#1_dense_photo.png}%
    & \includegraphics[width=0.133\linewidth]{figures/qualitative_comparison/Ours/#1_dense_c_gt.png}%
    & \includegraphics[width=0.133\linewidth]{figures/qualitative_comparison/Ours/#1_dense_c_pred.png}%
    & \includegraphics[width=0.133\linewidth]{figures/qualitative_comparison/C3Po/#1_dense_c_pred.png}%
    & \includegraphics[width=0.133\linewidth]{figures/qualitative_comparison/DUSt3R/#1_dense_c_pred.png}%
    & \includegraphics[width=0.133\linewidth]{figures/qualitative_comparison/LoFTR/#1_dense_c_pred.png}%
    & \includegraphics[width=0.133\linewidth]{figures/qualitative_comparison/combined_pose/#1_pose_pred_combined.png}
}

\definecolor{corrcol}{HTML}{E86301}
\definecolor{ourscol}{HTML}{8E63AE}
\definecolor{ourslightcol}{HTML}{C6AFDF}
\definecolor{c3pocol}{HTML}{F4B41A}
\definecolor{dust3rcol}{HTML}{B86A2B}
\definecolor{loftrcol}{HTML}{4DB6AC}

\newcommand{\camicon}[2]{%
    \tikz[baseline=-0.8ex, scale=0.6]{
        \draw[line width=0.8pt, draw=#1, fill=#2, line join=round] 
        (0,-0.08) -- (0.40,-0.18) -- (0.40,0.18) -- (0,0.08) -- cycle;
    }%
}

\newcommand{\paragraphdot}[1]{\paragraph{#1.}}
\newcommand{\methodName}{SceneAligner}

%% file: sec/0_abstract.tex
\begin{figure}[h!]
  \centering
  \includegraphics[width=\linewidth]{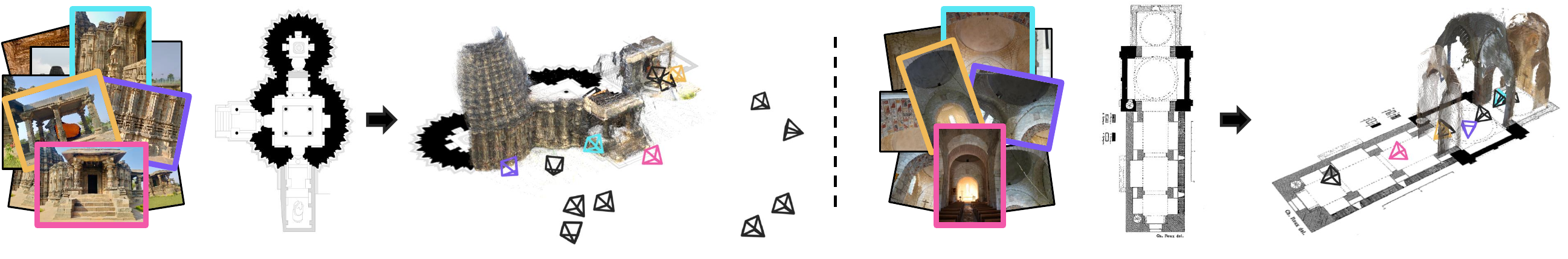}
  \vspace{-5mm}
  \caption{Given a collection of in-the-wild images and a rasterized floorplan, \emph{\methodName{}} reconstructs a gravity-aligned 3D point cloud from the images and globally aligns this reconstruction to the 2D floorplan map, thereby localizing the images within the floorplan. As illustrated above, our approach successfully aligns images capturing large-scale 3D environments, including exterior scenes (Doddabasappa Temple, left) and interior spaces (Église Saint-Martin d’Agonac, right).}
  \label{fig:teaser}
\end{figure}

\begin{abstract}
Many public buildings provide floorplans with a “you are here” indicator to help visitors orient themselves. Floorplan localization seeks to computationally replicate this capability by determining where visual observations were captured within a floorplan. However, existing methods typically assume controlled small-scale environments and precise vectorized floorplans, limiting their ability to operate in large-scale buildings and rasterized floorplans. In this work, we present an approach for performing floorplan localization in the wild by grounding the task in a reconstructed 3D representation of the scene. Given an unconstrained image collection, our method reconstructs a gravity-aligned 3D scene and projects it into a 2D density map that serves as a floorplan proxy. Floorplan localization is then formulated as aligning this proxy with the input floorplan via a 2D similarity transform. To bridge the appearance gap between density maps and architectural floorplans, we adapt a 2D foundation model to learn cross-modal correspondences, introducing a fine-tuning scheme that encourages semantically aligned matches while preserving structural consistency. Extensive experiments demonstrate substantial improvements over prior methods, including in extremely sparse settings with as little as a single input image. Our code and data will be publicly available.
\end{abstract}

%% file: sec/1_introduction.tex
\section{Introduction}
\label{sec:introduction}
Localizing camera observations within a provided 2D floorplan map is a fundamental task in 3D scene understanding, with applications in navigation, robotics, and augmented reality. Prior approaches~\cite{min2022laser,chen2024f3loc,grader2025supercharging} typically address this problem by exhaustively searching a discretized pose space, scoring candidate camera locations and orientations based on their consistency with a floorplan. This strategy inherently relies on access to precise, vectorized floorplans that encode fine-grained architectural primitives, such as exact wall layouts and openings. While effective in small-scale environments, these assumptions rapidly break down in large-scale, real-world settings, particularly in historic landmarks and monuments, where floorplans are often available only as rasterized or symbolic drawings and where architectural complexity far exceeds that of existing carefully-curated benchmarks. This raises a natural question: how can camera observations be localized within a floorplan in the \emph{wild}, when precise geometry is unavailable and exhaustive pose enumeration is no longer viable?

The recent emergence of 3D foundation models~\cite{VGGT,Pi3,liu2025worldmirroruniversal3dworld} has enabled accurate reconstruction of scene geometry directly from unconstrained image collections, even for large-scale environments captured under diverse viewpoints and illumination conditions. Given access to such high-fidelity geometric reconstructions, we argue that floorplan localization should be revisited from a fundamentally different perspective. To this end, we introduce \methodName{} (Figure~\ref{fig:teaser}), which reinterprets floorplan localization as a reconstruction and alignment problem. Rather than exhaustively enumerating and scoring camera poses over a discretized grid, our approach extracts a floorplan proxy from a reconstruction of the 3D scene, building upon prior floorplan reconstruction methods~\cite{RoomFormer,chen2019floorspinversecadfloorplans,FloorNet,PolyRoom} that recover 2D layouts from input 3D scans. Localization then reduces to globally aligning this proxy to the input floorplan via a 2D similarity transform. Specifically, we derive this proxy by orthographically projecting a gravity-aligned 3D reconstruction into a 2D density map.

To align this density map representation with the provided floorplan, we propose a feature matching learning scheme that estimates reliable cross-modal correspondences between the two modalities. While the density map provides a structurally-grounded proxy of the building layout, it differs significantly in appearance from architectural floorplans. To bridge this gap, we adapt a 2D foundation model (\emph{i.e.}, DINOv3~\cite{DINOv3}) to learn a shared feature space, introducing fine-tuning objectives that encourage semantically aligned cross-modal matches while preserving structural consistency. During inference, we extract a subset of reliable correspondences and estimate a 2D similarity transform that aligns the reconstructed 3D scene with the input floorplan, thereby enabling floorplan localization from unconstrained image collections.

We conduct extensive experiments comparing our approach against prior methods under both in-the-wild environments~\cite{C3Po} and synthetic indoor settings~\cite{Structured3D}. Our evaluation shows that \methodName{} achieves substantial performance improvements by factors ranging from two to three across most metrics on the in-the-wild testbed, while also outperforming indoor localization methods that rely on a discretized pose space. We further show that our approach remains effective for sparse image collections, surpassing baselines even when provided with a \emph{single} input view. Finally, we showcase the broader applicability of our approach by demonstrating that it enables the alignment of disjoint interior and exterior 3D reconstructions through registration to a shared floorplan.

%% file: sec/2_related_work.tex
\section{Related Work}
\label{sec:related_work}

\paragraphdot{Floorplan Localization}
Floorplan localization has been widely studied for indoor scene understanding, reconstruction, and navigation~\cite{boniardi2017robust,wang2015lost,chen2024f3loc,grader2025supercharging,UnLoc}. Early methods rely on depth-based cues from LiDAR~\cite{boniardi2017robust,boniardi2019pose,wang2019glfp,li2020online} or depth cameras~\cite{ito2014wrgdb}, often comparing extracted room edges to the 2D floorplan layouts while assuming known camera heights~\cite{boniardi2019robot,chu2015you}. More recent approaches embed images and floorplans into a shared feature space~\cite{howard2021lalaloc,howard2022lalaloc++}, or predict depth rays and probability volumes over the floorplan~\cite{chen2024f3loc,UnLoc}. To improve alignment and reduce ambiguities, several works incorporate semantic cues such as scene texts~\cite{wang2015lost}, CNN-extracted labels~\cite{mendez2020sedar}, pre-computed 3D maps~\cite{Kim2024_FullyGeometric}, or estimated semantic volumetric probabilities~\cite{grader2025supercharging}. However, these methods cannot address the challenge of localizing in-the-wild camera observations, where floorplans may be rasterized or symbolic drawings and input images come from unconstrained photo collections.

\paragraphdot{Learning Cross-Modal Correspondences}
Establishing correspondences is a fundamental problem in computer vision, underpinning tasks such as 3D reconstruction. Early methods rely on handcrafted descriptors~\cite{lowe2004distinctive,bay2006surf,rublee2011orb} with geometric verification~\cite{fischler1981random}, while recent approaches learn keypoints and matchers using deep visual features~\cite{detone2018superpoint,lindenberger2023lightglue,sarlin2020superglue}, enabling dense prediction (\eg, LoFTR~\cite{LoFTR}, DUSt3R~\cite{DUSt3R}). More recently, diffusion features~\cite{tang2023emergent} and self-supervised representations~\cite{DINOv3} have shown remarkable potential for correspondence estimation even across different visual domains~\cite{mikulinsky2025protosnap}. Nevertheless, matching natural images to symbolic representations remains challenging. C3Po~\cite{C3Po} learns correspondences between perspective photographs and symbolic 2D floorplans, but the extreme viewpoint and modality gap make this cross-modal matching highly under-constrained, as it requires reasoning about the underlying 3D geometry connecting perspective views to top-down layouts. In contrast, our method bridges this gap via an intermediate density map derived from 3D scene reconstruction, naturally connecting unconstrained photographs to abstract floorplans.

\paragraphdot{3D Scene Understanding from Internet Photo Collections}
Prior work has explored spatial understanding of large-scale environments by analyzing visual patterns, viewpoints, or metadata~\cite{weyand2013discovering,simon2007scene,simon2008scene,russell20133d,wu2021towers,li2026longtailinternetphotoreconstruction}. Notable efforts include assembling disjoint 3D indoor reconstructions using annotated maps and crowd flow~\cite{martin20143d}, aligning interior and exterior 3D scenes via scene semantics~\cite{cohen2016indoor}, and registering photo collections to a 3D reference model using semantic features~\cite{cohen2026scene}. However, prior work cannot directly align unconstrained photos with floorplans. Recent advances in 3D foundation models~\cite{VGGT,Pi3,liu2025worldmirroruniversal3dworld} and gravity estimator~\cite{GeoCalib} make gravity-aligned reconstruction feasible, which we leverage to revisit floorplan localization through 3D-grounded scene understanding.

%% file: sec/3_method.tex
\begin{figure}[t!]
  \centering
  \includegraphics[width=\linewidth]{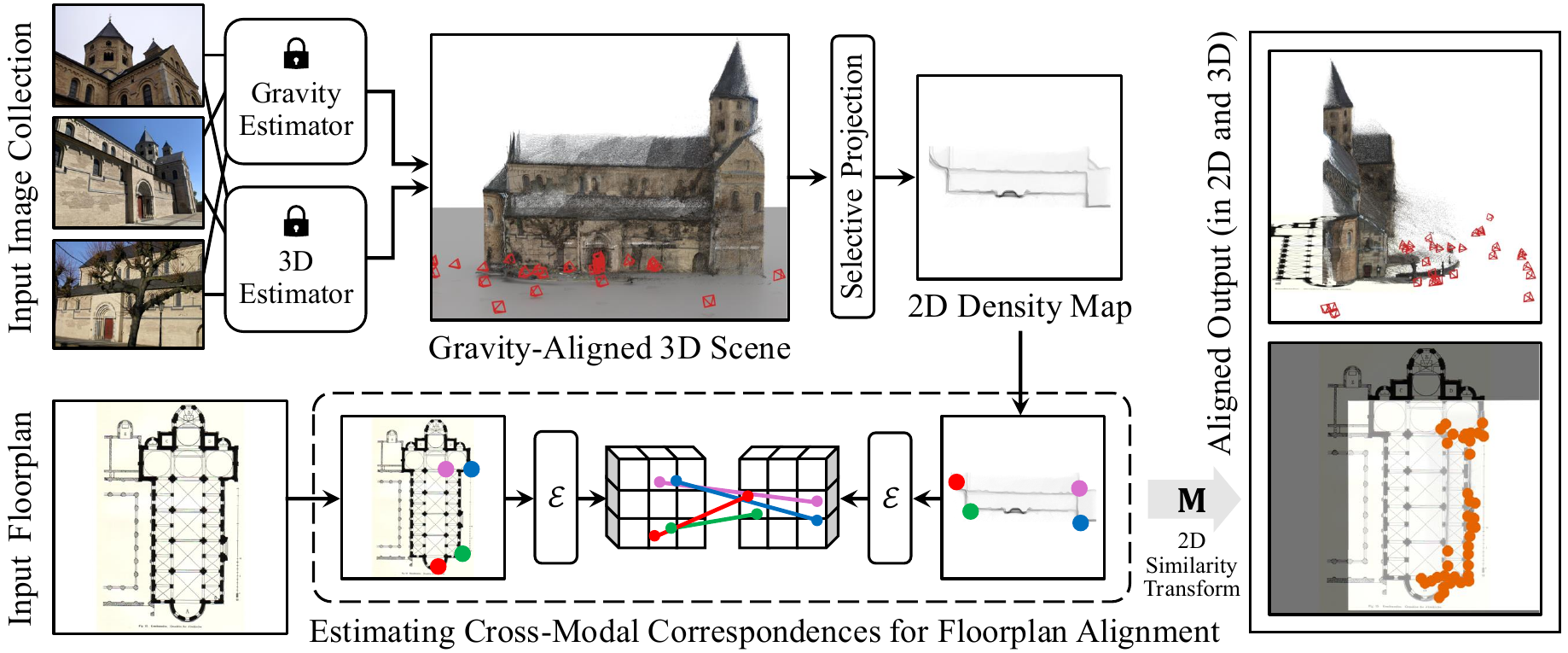}
  \vspace{-4.25mm}
  \caption{\textbf{\methodName{}.} Given in-the-wild images and a floorplan, it reconstructs a gravity-aligned 3D scene, extracts a 2D density map via projection, and solves for a 2D similarity transform $\mathbf{M}$ via correspondence estimation between the density map and floorplan using a shared encoder $\mathcal{E}$. Reliable correspondences used to compute $\mathbf{M}$ are overlaid (in \textcolor{corrcol}{orange}) on the aligned density map.}
  \label{fig:method_overview}
  \vspace{-0.5mm}
\end{figure}

\section{Method}
\label{sec:method}
As illustrated in Figure~\ref{fig:method_overview}, our method recovers a gravity-aligned 3D scene (Sec.~\ref{sec:gravity_aligned_3d_scene_reconstruction}), derives a floorplan proxy  (Sec.~\ref{sec:density_map_extraction}), and predicts a similarity transform for floorplan alignment by estimating correspondences between the floorplan and proxy (Sec.~\ref{sec:correspondence_estimation}). We describe each step below.

\subsection{Reconstructing a Gravity-Aligned 3D Scene}
\label{sec:gravity_aligned_3d_scene_reconstruction}
Formulating floorplan localization as a 3D reconstruction and alignment problem, we first recover the scene geometry and camera poses from an unconstrained image collection $\mathcal{I} = \{I_{1}, \dots, I_{N}\}$, where each image ${I}_{i}$ has a resolution of $H_{i} \times W_{i}$. We leverage a 3D foundation model (\eg, $\pi^{3}$~\cite{Pi3}, VGGT~\cite{VGGT}) to estimate 3D points $\mathcal{P}^{\mathrm{c}}_{i} \in \mathbb{R}^{H_{i}W_{i} \times 3}$ in a camera coordinate frame, along with relative camera poses that map each camera frame to the reference frame of $I_{1}$.

To align the 3D scene geometry with the physical ground plane, we predict a gravity direction per image using GeoCalib~\cite{GeoCalib}, transform each gravity vector into the reference frame using the corresponding relative camera pose, and select their medoid as a robust gravity estimate $\mathbf{g}$. We then compute a rigid transformation via Gram-Schmidt orthogonalization to align $\mathbf{g}$ with the vertical $y$-axis. By applying this transformation, we obtain gravity-aligned 3D points $\mathcal{P}^{\mathrm{g}}_{i} \in \mathbb{R}^{H_{i}W_{i} \times 3}$. This ensures that the reconstructed 3D geometry and the input floorplan share a common horizontal plane.

\subsection{Extracting a 2D Density Map as a Floorplan Proxy}
\label{sec:density_map_extraction}
With the gravity-aligned 3D scene, we can extract a 2D density map via orthographic projection. However, directly projecting all points $\mathcal{P}^{\mathrm{g}}_{*} = \bigcup_{i=1}^{N} \mathcal{P}^{\mathrm{g}}_{i}$ is vulnerable to outliers such as faraway backgrounds or sky regions. Unlike the clean density maps assumed in the floorplan reconstruction literature~\cite{RoomFormer,chen2019floorspinversecadfloorplans,FloorNet,PolyRoom}, these artifacts introduce significant noise into the resulting density map, making the subsequent estimation of the 2D similarity transform (Sec.~\ref{sec:correspondence_estimation}) unstable.

To obtain a clean and structurally meaningful density map, we identify a subset of 3D points $\mathcal{\bar{P}}^{\mathrm{g}}_{*}$ that are (i) \emph{geometrically reliable}, (ii) \emph{spatially bounded}, and (iii) \emph{representative of vertical structures}. Specifically, we first remove unreliable geometry based on the 3D reconstruction model's confidence scores. Next, we discard horizontal outliers to retain points within the spatial extent of the scene. Finally, as floorplans primarily depict vertical structures, we filter points along the gravity-aligned axis to suppress floor and ceiling surfaces while preserving layout-defining geometry such as walls.

After filtering, the remaining points $\mathcal{\bar{P}}^{\mathrm{g}}_{*}$ are orthographically projected onto the horizontal $xz$-plane, where we count the number of points falling into each grid cell. We then apply gamma correction and normalization to ensure consistent visibility across scenes, yielding a density map $D \in \mathbb{R}^{H \times W \times 1}$ with a top-down, line-drawing modality similar to the reference floorplan $F \in \mathbb{R}^{H \times W \times 3}$.

\subsection{Learning Cross-Modal Floorplan--Density Map Correspondences}
\label{sec:correspondence_estimation}
Equipped with the extracted density map, we perform floorplan alignment by estimating a 2D similarity transform $\mathbf{M} \in \mathrm{Sim}(2)$, which serves to align the reconstructed scene with the floorplan. This transform is parameterized by a scale $s \in \mathbb{R}^{+}$, rotation $\mathbf{R} \in \mathrm{SO}(2)$, and translation $\mathbf{t} \in \mathbb{R}^{2}$. We estimate $\mathbf{M}$ via RANSAC~\cite{fischler1981random} using correspondences between the density map and floorplan.

\begin{figure}[t!]
  \centering
  \includegraphics[width=\linewidth]{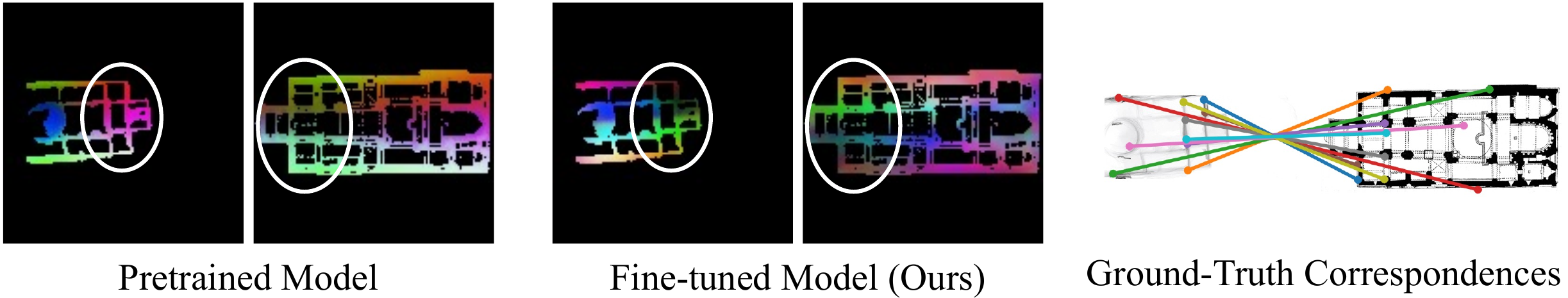}
  \vspace{-3.5mm}
  \caption{\textbf{Adapting a 2D foundation model for floorplan alignment.} We provide PCA visualizations of features before and after our fine-tuning scheme. As illustrated above, the pretrained DINOv3~\cite{DINOv3} struggles to bridge the appearance gap, \emph{e.g.}, corresponding regions (white circles) map to different RGB colors. By contrast, our fine-tuning significantly refines the semantic cross-modal alignment. For reference, we show randomly-sampled ground-truth correspondences on the right.} 
  \label{fig:pca_visualizations}
  \vspace{-2.5mm}
\end{figure}

However, establishing cross-modal correspondences is non-trivial. As demonstrated by the PCA visualizations in Figure~\ref{fig:pca_visualizations}, even a 2D foundation model~\cite{DINOv3} fails to produce semantically aligned features due to the severe appearance gap between the noisy density map and the clean architectural drawing. To address this, we propose a fine-tuning scheme that facilitates \textit{semantic alignment} and enforces \textit{structural consistency} among correspondences, enabling robust similarity transform estimation. We adopt DINOv3~\cite{DINOv3} as a shared encoder $\mathcal{E}$, freeze its pretrained weights, and inject trainable Low-Rank Adaptation (LoRA)~\cite{LoRA} layers. These layers are optimized via:
\begingroup
\setlength{\abovedisplayskip}{4.5pt}
\setlength{\belowdisplayskip}{4.5pt}
    \begin{align}
        \mathcal{L} = 
        \lambda_{\mathrm{feat}}\mathcal{L}_{\mathrm{feat}} + 
        \lambda_{\mathrm{regr}}\mathcal{L}_{\mathrm{regr}} + 
        \lambda_{\mathrm{topo}}\mathcal{L}_{\mathrm{topo}} +
        \lambda_{\mathrm{geo}}\mathcal{L}_{\mathrm{geo}},
    \end{align}
\endgroup
where $\lambda_{\mathrm{feat}}, \lambda_{\mathrm{regr}}, \lambda_{\mathrm{topo}}, \lambda_{\mathrm{geo}}$ are loss coefficients. 

\paragraphdot{Feature Matching Objective}
To establish cross-modal correspondences, we employ a contrastive feature matching loss $\mathcal{L}_{\mathrm{feat}}$. The encoder $\mathcal{E}$ extracts feature maps $\mathbf{F}^{D}, \mathbf{F}^{F} \in \mathbb{R}^{H' \times W' \times C}$ from the density map $D$ and floorplan $F$, where $H' = H/16$ and $W' = W/16$. During training, we sample $Q$ ground-truth correspondence pairs $\{ (\mathbf{p}_{i}^{D}, \mathbf{p}_{i}^{F}) \}^{Q}_{i=1}$ and compute their feature vectors $\mathbf{f}_{i}^{D}, \mathbf{f}_{i}^{F} \in \mathbb{R}^{C}$ via bilinear interpolation and $\ell_{2}$-normalization. We then compute a similarity matrix $\mathbf{S} \in \mathbb{R}^{Q \times Q}$ where $\mathbf{S}_{ij} = \langle \mathbf{f}_{i}^{D}, \mathbf{f}_{j}^{F} \rangle$ measures the cosine similarity, and minimize a symmetric InfoNCE loss~\cite{InfoNCE}:
\begingroup
\setlength{\abovedisplayskip}{4.5pt}
\setlength{\belowdisplayskip}{4.5pt}
    \begin{align}
    \label{eq:contrastive_loss}
        \mathcal{L}_{\mathrm{feat}} = - \frac{1}{2Q} \sum_{i=1}^{Q} \left[ \log \left( \frac{\exp(\mathbf{S}_{ii}/\tau)}{\sum_{k=1}^{Q} \exp(\mathbf{S}_{ik}/\tau)} \right) + \log \left( \frac{\exp(\mathbf{S}_{ii}/\tau)}{\sum_{k=1}^{Q} \exp(\mathbf{S}_{ki}/\tau)} \right) \right],
    \end{align}
\endgroup
where $\tau$ is a temperature scaling parameter.

\paragraphdot{Coordinate Regression Objective}
Existing correspondence estimation approaches typically select the maximum similarity on a patch-level feature map and assign its centroid as the match, leading to quantization errors (\eg, $\pm$ 8 pixels for a $16 \times 16$ patch size). To achieve sub-patch precision, we introduce a coordinate regression loss $\mathcal{L}_{\mathrm{regr}}$ using a differentiable soft-argmax.

Given the density map feature vectors $\{ \mathbf{f}_{i}^{D} \}_{i=1}^{Q}$ and flattened floorplan features $\mathbf{\bar{F}}^{F} \in \mathbb{R}^{H'W' \times C}$, we compute a similarity matrix $\mathbf{S}' \in \mathbb{R}^{Q \times H'W'}$, convert it into a spatial probability distribution over floorplan patches via softmax, and estimate floorplan coordinates $\{ \mathbf{\hat{p}}_{i}^{F} \}_{i=1}^{Q}$ as the expectation over patch centroids. We supervise the prediction with a confidence-weighted Huber loss $\mathcal{H}_{\delta}(\cdot)$ between the predicted $\mathbf{\hat{p}}_{i}^{F}$ and ground-truth $\mathbf{p}_{i}^{F}$ floorplan coordinates:
\begingroup
\setlength{\abovedisplayskip}{4.5pt}
\setlength{\belowdisplayskip}{4.5pt}
    \begin{align}
    \mathcal{L}_{\mathrm{regr}} = \frac{\sum_{i=1}^{Q} w_{i} \mathcal{H}_\delta (\| \mathbf{\hat{p}}_{i}^{F} - \mathbf{p}_{i}^{F} \|_{2})}{\sum_{i=1}^{Q} w_{i}},
    \end{align}
\endgroup
where $w_{i}$ is the maximum softmax probability for the $i$-th correspondence, serving as a confidence weight that reflects the sharpness of the distribution.

\paragraphdot{Structural Consistency Regularization}
Point-wise objectives can lead to degenerate similarity transforms when the spatial structure of correspondences collapses. To prevent this, we introduce a topology preservation loss $\mathcal{L}_{\mathrm{topo}}$ and a geometry consistency loss $\mathcal{L}_{\mathrm{geo}}$ as self-supervised structural priors, leveraging the fact that a similarity transform preserves relative angles and distance ratios.

$\mathcal{L}_{\mathrm{topo}}$ enforces angular consistency on triplets $(i, j, k)$ sampled from the $Q$ correspondences via:
\begingroup
\setlength{\abovedisplayskip}{4.5pt}
\setlength{\belowdisplayskip}{4.5pt}
    \begin{align}
    \label{eq:angle_consistency_loss}
        \mathcal{L}_{\mathrm{topo}} = \frac{\sum_{(i,j,k)} w_{i} w_{j}  w_{k} \left[\mathcal{H}_\delta(\cos \hat{\theta}_{ijk}^{F} \!-\! \cos \theta_{ijk}^{D}) \!+\! \mathcal{H}_\delta(\sin \hat{\theta}_{ijk}^{F} \!-\! \sin \theta_{ijk}^{D}) \right]}{\sum_{(i,j,k)} w_{i} w_{j} w_{k}},
    \end{align}
\endgroup
where $\hat{\theta}_{ijk}^{F}$ and $\theta_{ijk}^{D}$ are the corresponding angles of triangles by $\{ \mathbf{\hat{p}}_{i}^{F}, \mathbf{\hat{p}}_{j}^{F}, \mathbf{\hat{p}}_{k}^{F} \}$ and $\{ \mathbf{p}_{i}^{D}, \mathbf{p}_{j}^{D}, \mathbf{p}_{k}^{D} \}$.

$\mathcal{L}_{\mathrm{geo}}$ enforces consistent distance ratios over sampled pairs $(i, j)$ by penalizing deviations of the log-distance ratio $\Delta_{ij} =  \log \left( \| \mathbf{\hat{p}}_{i}^{F} - \mathbf{\hat{p}}_{j}^{F} \|_2 / \| \mathbf{p}_{i}^{D} - \mathbf{p}_{j}^{D} \|_2 \right)$ from the weighted mean $\bar{\Delta}$ via:
\begingroup
\setlength{\abovedisplayskip}{4.5pt}
\setlength{\belowdisplayskip}{4.5pt}
    \begin{align}
    \label{eq:spatial_consistency_loss}
    \mathcal{L}_{\mathrm{geo}} = \frac{\sum_{(i,j)} w_{i} w_{j} \mathcal{H}_\delta (\Delta_{ij} - \operatorname{sg}(\bar{\Delta}))}{\sum_{(i,j)} w_{i} w_{j}},
    \end{align}
\endgroup
where $\operatorname{sg}(\cdot)$ denotes the stop-gradient operator.

\paragraphdot{Floorplan Alignment from Reliable Correspondences} 
To robustly estimate the 2D similarity transform $\mathbf{M}$, we identify \emph{highly-confident} and \emph{mutually-close} correspondences. We retain the top 50\% ranked by confidence $w_{i}$ and apply mutual nearest neighbor (MNN) matching. The resulting reliable correspondences are used to estimate $\mathbf{M}$ via RANSAC~\cite{fischler1981random}. We then apply $\mathbf{M}$ to the 3D points $\mathcal{P}^{\mathrm{g}}_{*}$ by transforming their horizontal $(x,z)$ coordinates while scaling the vertical $y$ coordinates by the scale $s$ to maintain structural proportions. Camera poses are transformed accordingly, producing a floorplan-aligned 3D scene for accurate floorplan localization in the wild.

%% file: sec/4_experiments.tex
\section{Experiments}
\label{sec:experiments}
We conduct comprehensive experiments to evaluate our method under in-the-wild environments as well as synthetic indoor settings. In Section~\ref{sec:experimental_setup}, we describe our experimental setup such as implementation details. In Section~\ref{sec:evaluation_on_in_the_wild_data}, we evaluate the proposed approach on in-the-wild data, including comparisons with baselines (Sec.~\ref{sec:evaluation_in_the_wild_comparison_w_corres_methods}) and robustness analysis under sparse-view settings (Sec.~\ref{sec:evaluation_in_the_wild_robustness_analysis}). In Section~\ref{sec:evaluation_on_synthetic_data}, we evaluate our method on a synthetic indoor dataset for comparison with prior floorplan localization approaches. In Section~\ref{sec:applications}, we showcase downstream applications such as interior-exterior 3D scene alignment using a reference floorplan.

The appendix supplements our main results with additional experiments. For example, we analyze the stability of our model against various hyperparameters (Sec.~\ref{sec:supp_robustness_to_density_map_hyperparameters}) and validate our design choices, including the ablation study on learning objectives (Table~\ref{table:loss_ablation}), correspondence filtering strategies (Table~\ref{table:effect_of_correspondence_filtering}), LoRA configurations (Table~\ref{table:supp_lora_hyperparameters}), and 3D reconstruction models (Table~\ref{table:supp_effect_of_3d_recon_models}). We also provide an HTML viewer that shows 360$^{\circ}$ view comparisons of floorplan-aligned 3D scenes.

\begin{table*}[!t]
    \centering
    \caption{\textbf{Camera pose estimation on C3~\cite{C3Po}.} Our fine-tuning scheme significantly improves the 2D foundation model~\cite{DINOv3}, outperforming prior methods by a large margin across all metrics.}
    \label{table:camera_pose_image_floorplan}
    \vspace{-1.5mm}
    \renewcommand{\arraystretch}{1.1}
    \setlength{\tabcolsep}{5pt}
    \resizebox{\linewidth}{!}{
        \begin{tabular}{l cccc c ccc c c}
        
        \noalign{\hrule height 0.85pt}
        & \multicolumn{4}{c}{Angular Recall\,$\uparrow$} & \!\!\! & \multicolumn{3}{c}{Positional Recall\,$\uparrow$} & \!\!\! & \!\!\!Ang.\,\&\,Pos.\,$\uparrow$\!\!\!
        \\ 
        
        \cline{2-5}
        \cline{7-9}
        \cline{11-11}
        Method & \,@\,$5^\circ$ & \,@\,$10^\circ$ & \,@\,$20^\circ$ & \,@\,$30^\circ$ & \!\!\! & \,@\,$5\%$ & \,@\,$10\%$ & \,@\,$20\%$ & \!\!\! & \!\!\,@\,($30^\circ,20\%$)\!\!
        \\
        
        \noalign{\hrule height 0.7pt}
        LoFTR~\cite{LoFTR} & 4.67 & 8.31 & 15.11 & 22.80 & \!\!\! & 1.09 & 4.27 & 13.96 & \!\!\! & 3.07 
        \\
        DUSt3R~\cite{DUSt3R} & 5.14 & 10.06 & 18.02 & 24.59 & \!\!\! & 1.54 & 5.15 & 14.71 & \!\!\! & 4.64 
        \\
        C3Po~\cite{C3Po} & 24.45 & 36.90 & 48.59 & 54.89 & \!\!\! & 16.60 & 28.59 & 42.12 & \!\!\! & 32.96 
        \\ 
        \midrule
        DINOv3~\cite{DINOv3} & 14.25 & 19.59 & 28.87 & 35.33 & \!\!\! & 6.75 & 14.23 & 26.67 & \!\!\! & 18.28
        \\
        \cellcolor{gray!7.5}\textbf{Ours} & \cellcolor{gray!7.5}\textbf{65.91} & \cellcolor{gray!7.5}\textbf{75.93} & \cellcolor{gray!7.5}\textbf{80.23} & \cellcolor{gray!7.5}\textbf{83.08} & \cellcolor{gray!7.5}\!\!\! & \cellcolor{gray!7.5}\textbf{50.87} & \cellcolor{gray!7.5}\textbf{65.58} & \cellcolor{gray!7.5}\textbf{76.08} & \cellcolor{gray!7.5}\!\!\! & \cellcolor{gray!7.5}\textbf{73.58} 
        \\
        \noalign{\hrule height 0.85pt}
    \end{tabular}}
\end{table*}

\begin{table*}[!t]
    \centering
    \caption{\textbf{Correspondence estimation on C3~\cite{C3Po}.} Our adaptation of the 2D foundation model~\cite{DINOv3} achieves substantially higher accuracy than prior methods across all pixel-level metrics.}
    \label{table:correspondence_image_floorplan}
    \vspace{-1.5mm}
    \renewcommand{\arraystretch}{1.1}
    \setlength{\tabcolsep}{7pt}
    \resizebox{\linewidth}{!}{
        \begin{tabular}{l cccccc c}
        
        \noalign{\hrule height 0.85pt} 
        \! & \multicolumn{6}{c}{PCK\,$\uparrow$} & 
        \\ 
        
        \cline{2-7}
        \!Method & \,@\,1\%\, & \,@\,3\%\, & \,@\,5\%\, & \,@\,10\%\, & \,@\,15\%\, & \,@\,30\%\, & \,RMSE\,$\downarrow$\,
        \\
        
        \noalign{\hrule height 0.7pt}
        \!LoFTR~\cite{LoFTR} & 0.14 & 1.04 & 2.49 & 7.40 & 13.59 & 37.13 & 0.3227 
        \\
        \!DUSt3R~\cite{DUSt3R} & 0.09 & 0.76 & 2.10 & 7.69 & 15.78 & 40.20 & 0.2844 
        \\
        \!C3Po~\cite{C3Po} & 9.06 & 28.37 & 34.91 & 45.83 & 54.26 & 72.25 & 0.1780 
        \\ 
        \midrule
        \!DINOv3~\cite{DINOv3} & 0.73 & 3.95 & 7.95 & 16.25 & 24.31 & 47.01 & 0.2624 
        \\
        \cellcolor{gray!7.5}\!\textbf{Ours} &  \cellcolor{gray!7.5}\textbf{20.42} & \cellcolor{gray!7.5}\textbf{58.13} & \cellcolor{gray!7.5}\textbf{69.87} & \cellcolor{gray!7.5}\textbf{79.94} & \cellcolor{gray!7.5}\textbf{83.64} & \cellcolor{gray!7.5}\textbf{90.45} & \cellcolor{gray!7.5}\textbf{0.0776} 
        \\
        \noalign{\hrule height 0.85pt}
    \end{tabular}}
\end{table*}

\subsection{Experimental Setup}
\label{sec:experimental_setup}

\paragraphdot{Implementation Details}
We reconstruct 3D scene geometry using $\pi^{3}$~\cite{Pi3} and predict gravity using GeoCalib~\cite{GeoCalib}. For correspondence estimation, we employ a pretrained DINOv3 ViT-B/16~\cite{DINOv3} and inject trainable LoRA~\cite{LoRA} layers, which are optimized using AdamW~\cite{AdamW} with a learning rate of $10^{-4}$. We adopt the same settings (\eg, model hyperparameters) for both in-the-wild and synthetic indoor evaluations. Comprehensive details are provided in the appendix (Sec.~\ref{sec:supp_implementation_details}).

\paragraphdot{Varying-View Inference}
We evaluate our method across varying numbers of input views. By default, scenes with $N > 150$ images are partitioned into $\lceil N / 150 \rceil$ chunks, whereas smaller scenes are processed entirely. We also evaluate under sparser settings, \eg, Ours ($= 1$) takes a single image as input for 3D reconstruction with other settings unchanged in Figures~\ref{fig:performance_across_varying_view_counts} and~\ref{fig:visualize_sparse_recon}.

\begin{figure*}[t]
    \centering
    \setlength{\tabcolsep}{1.75pt} 
    
    \begin{tabular}{@{} c c c c c c<{\hspace{\tabcolsep}} : >{\hspace{\tabcolsep}}c @{}}
        
        & \textcolor{red}{GT} & 
        \textcolor{ourscol}{Ours} & 
        \textcolor{c3pocol}{C3Po} & 
        \textcolor{dust3rcol}{DUSt3R} & 
        \textcolor{loftrcol}{LoFTR} & \\[-2pt]

        \comparowseven{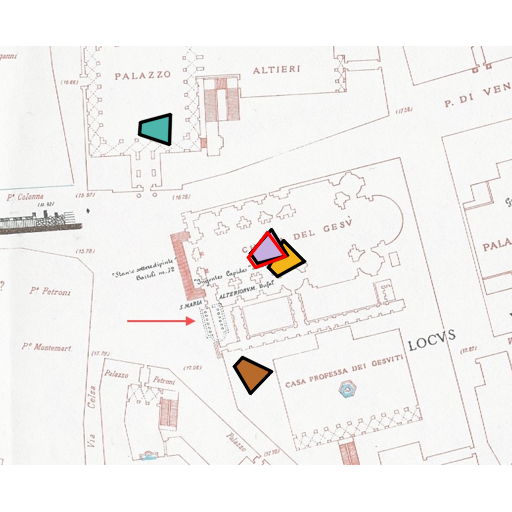} \\[-6pt]
        \comparowseven{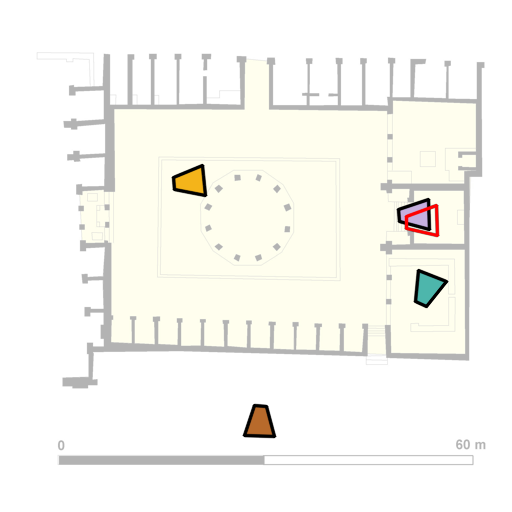} \\[-6pt]
        \comparowseven{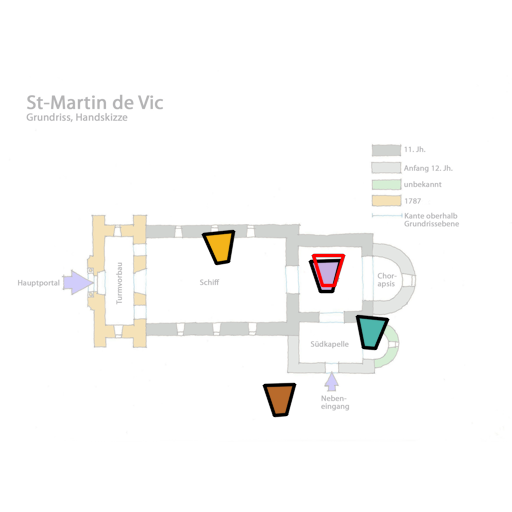} \\[-6pt]
        \comparowseven{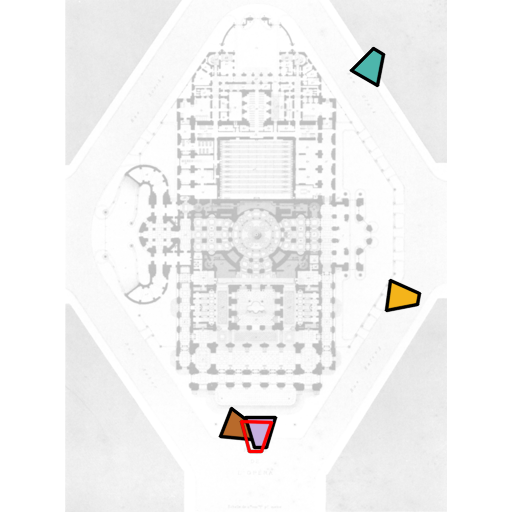}
    \end{tabular}

    \vspace{-1mm}
    \caption{\textbf{Qualitative comparison in the wild.} We compare correspondence predictions across baselines and our method, with floorplan localization results shown on the right. Cameras are illustrated in corresponding colors (\eg, {\protect\camicon{red}{white}} for \textcolor{red}{GT}, and {\protect\camicon{black}{ourslightcol}} for \textcolor{ourscol}{Ours}).}
    \vspace{-1.5mm}
    \label{fig:qualitative_comparison}
\end{figure*}

\subsection{Evaluation on In-the-Wild Data}
\label{sec:evaluation_on_in_the_wild_data}

We evaluate our method at multiple levels of granularity, reporting both image-level camera pose estimation and pixel-level correspondence metrics.

\paragraphdot{C3 Dataset}
C3~\cite{C3Po} is a large-scale in-the-wild dataset of diverse photographs paired with floorplans, providing camera pose and correspondence annotations. These annotations are derived by registering Structure-from-Motion reconstructions to floorplans, inevitably including geometric misalignments in unconstrained settings. To ensure a reliable testbed, we curate a clean subset by pruning samples with severe errors. For example, images with severe optical distortions or non-photographic content are removed (Figure~\ref{fig:supp_bad_samples}). Importantly, this filtering preserves the original dataset's diversity, retaining $574$ out of $597$ scenes (96.15\%). Details are provided in the appendix (Sec.~\ref{sec:supp_dataset_statistics}).

\paragraphdot{Camera Pose Evaluation Metrics}
Camera pose estimation is measured using \textit{Angular Recall} and \textit{Positional Recall}, which evaluate camera yaw errors and 2D horizontal center distances against the ground truth, respectively. Following C3Po~\cite{C3Po}, we report Angular Recall at $\{5^{\circ}, 10^{\circ}, 20^{\circ}, 30^{\circ}\}$, Positional Recall at $\{5\%, 10\%, 20\%\}$ of the floorplan diagonal length, and their combined recall.

\paragraphdot{Correspondence Evaluation Metrics}
Pixel-level correspondence is measured using \textit{Percentage of Correct Keypoints (PCK)} and \textit{Root Mean Square Error (RMSE)}. PCK quantifies the proportion of correspondences within a distance threshold from the ground truth. Following C3Po~\cite{C3Po}, distances are normalized by the floorplan resolution. We report PCK at $\{1\%, 3\%, 5\%, 10\%, 15\%, 30\%\}$.

\subsubsection{Comparison with Correspondence-based Methods}
\label{sec:evaluation_in_the_wild_comparison_w_corres_methods}

\paragraphdot{Baselines}
We compare against correspondence-based methods~\cite{C3Po,DUSt3R,LoFTR} on the clean subset of C3. C3Po~\cite{C3Po} builds on the DUSt3R architecture~\cite{DUSt3R} and is fine-tuned on the C3 dataset to learn correspondences between perspective photographs and abstract floorplans. We follow C3Po's evaluation protocol for these methods: candidate camera poses are predicted by solving epipolar geometry from estimated correspondences, where the candidate closest to the ground truth is selected.

To isolate the contribution of our adaptation scheme, we also compare against pretrained DINOv3~\cite{DINOv3} as the encoder $\mathcal{E}$ in our inference pipeline. Additional ablations are provided in the appendix (Sec.~\ref{sec:supp_ablation_study}).

\paragraphdot{Quantitative Evaluation}
As shown in Tables~\ref{table:camera_pose_image_floorplan} and~\ref{table:correspondence_image_floorplan}, our method significantly outperforms all baselines at both image-level and pixel-level, achieving over 100\% improvements across most evaluation metrics. For example, in terms of combined angular-positional recall, our approach surpasses the strongest baseline, C3Po, by 123.24\% (73.58 vs. 32.96) and the pretrained DINOv3 baseline by 302.52\% (73.58 vs. 18.28). With regard to RMSE, C3Po's error is 129.38\% (0.1780 vs. 0.0776) higher than ours, and DINOv3's error is 238.14\% (0.2624 vs. 0.0776) higher. The large gap over the DINOv3 baseline highlights the contribution of the proposed fine-tuning scheme. These results demonstrate the effectiveness of our approach for floorplan localization in the wild.

\paragraphdot{Qualitative Evaluation}
Figure~\ref{fig:qualitative_comparison} visualizes correspondence and camera pose estimation results against correspondence-based baselines, showing that our predictions align most closely with the ground truth. In the second row, our estimated correspondences structurally match the ground truth, whereas all baselines produce physically implausible results. This highlights the advantage of our 3D-grounded approach that enforces spatial rigidity through gravity-aligned 3D scene reconstruction. Furthermore, as shown in the third row, our method successfully localizes minimal-context photos (\eg, wall drawings) by leveraging global 3D scene reconstructions, addressing a severely challenging scenario where single-view estimation methods typically fail.

\begin{figure}[t!]
  \centering
  \includegraphics[width=\linewidth]{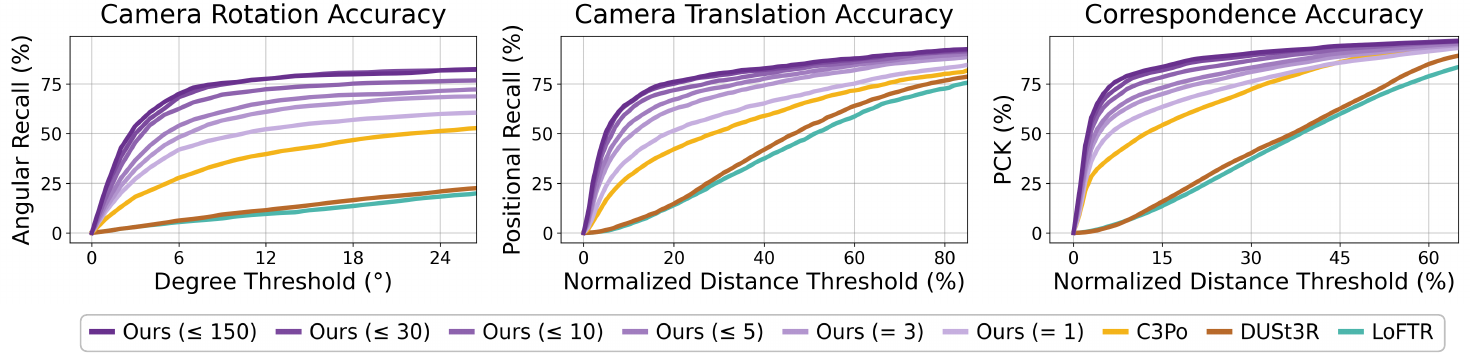}
  \vspace{-3.5mm}
  \caption{\textbf{Performance across varying view counts on C3~\cite{C3Po}.} We evaluate the proposed method using different numbers of input images for 3D reconstruction (\eg, $\leq150$ denotes a maximum of $150$ images per reconstruction). Notably, Ours ($= 1$) already outperforms C3Po~\cite{C3Po} by a large margin, while Ours ($\leq30$) is on par with Ours ($\leq150$).} 
  \label{fig:performance_across_varying_view_counts}
  \vspace{-1mm}
\end{figure}

\begin{figure}[t!]
  \centering
  \includegraphics[width=\linewidth]{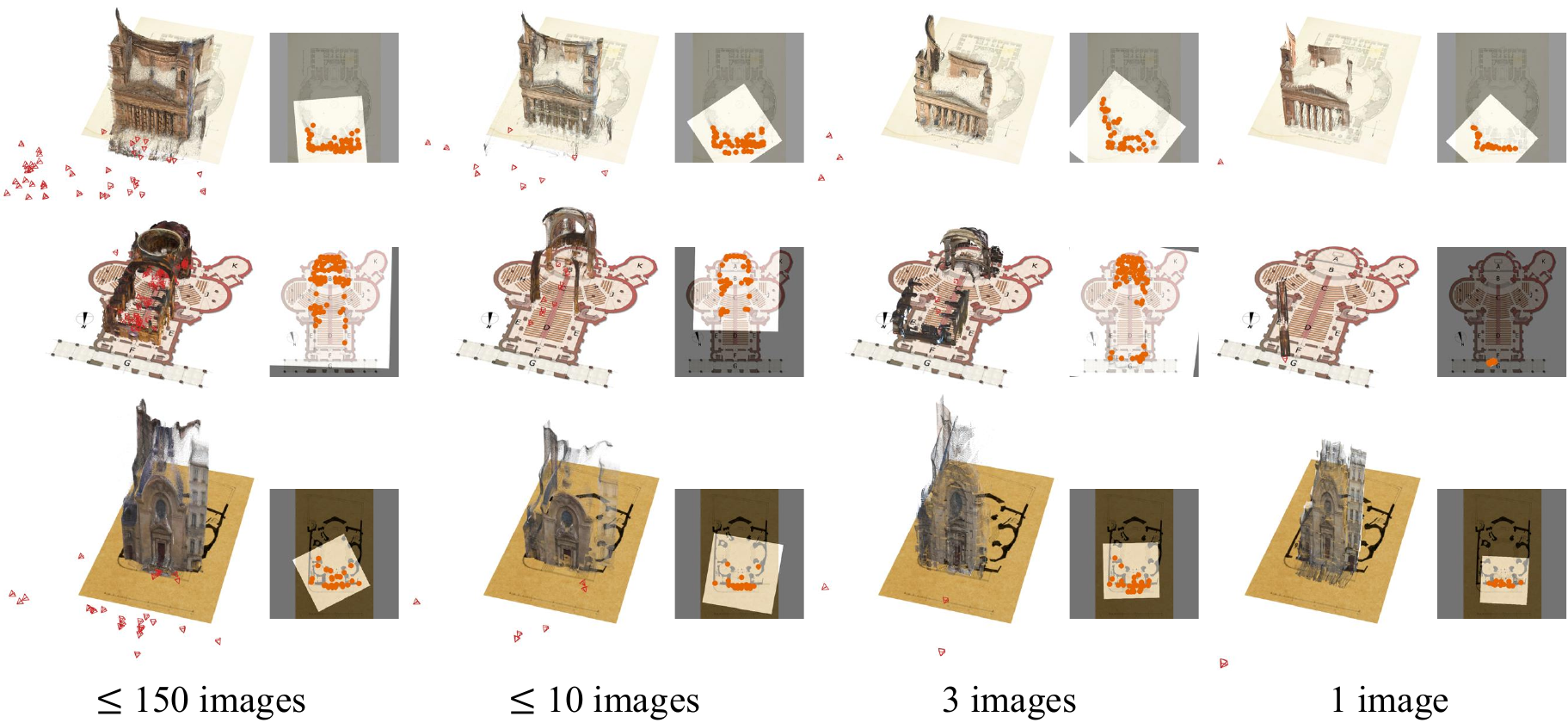}
  \vspace{-3.5mm}
  \caption{\textbf{In-the-wild floorplan alignment across varying view counts.} We visualize 3D points used for density map extraction and resulting density maps with reliable correspondences overlaid (in \textcolor{corrcol}{orange}).
  Sparse views often suffice to recover geometry informative enough for floorplan alignment.}
  \label{fig:visualize_sparse_recon}
\end{figure}

\subsubsection{Robustness to Sparse Inputs}
\label{sec:evaluation_in_the_wild_robustness_analysis} 
We evaluate how effectively our method localizes given a limited number of input images. As shown in Figure~\ref{fig:performance_across_varying_view_counts}, the proposed method significantly outperforms C3Po even in the single-view setting, and achieves accuracy comparable to the 150-view setting with as few as 10 to 30 views. Figure~\ref{fig:visualize_sparse_recon} further shows that even a small set of images is typically sufficient to construct a structurally meaningful density map, enabling accurate floorplan alignment. However, the single-view setting sometimes suffers from localization ambiguity, as shown in the second row. Because a single image captures only a limited region of the scene, its reconstructed geometry (\eg, a single wall segment) may match multiple similar structures on the floorplan, leading to the accuracy drop observed in Figure~\ref{fig:performance_across_varying_view_counts}.

\subsection{Evaluation on Synthetic Data}
\label{sec:evaluation_on_synthetic_data}
In addition to in-the-wild scenarios, we also evaluate our 3D-grounded approach on a standard indoor benchmark~\cite{Structured3D}. This enables direct comparison with prior localization methods.

\paragraphdot{Structured3D Dataset}
Structured3D~\cite{Structured3D} is a large-scale synthetic indoor dataset of photorealistic renderings paired with vectorized floorplans, providing precise geometry under controlled environments. This dataset has been widely used as a testbed for floorplan localization.

\paragraphdot{Camera Pose Evaluation Metrics}
Camera pose estimation is measured using \textit{Angular Recall} and \textit{Positional Recall}. Following the standard protocol~\cite{min2022laser,chen2024f3loc,UnLoc}, we report Positional Recall at $\{ 0.1\mathrm{m}, 0.5\mathrm{m}, 1\mathrm{m} \}$ and the combined angular-positional recall at $(30^{\circ}, 1\mathrm{m})$.

\subsubsection{Comparison with Indoor Localization Methods}
\label{sec:evaluation_synthetic_comparison_w_indoor_localization_methods}

\paragraphdot{Baselines}
We compare against state-of-the-art indoor localization methods~\cite{min2022laser,chen2024f3loc,UnLoc}, which rely on pre-defined discretized pose spaces, \eg, an $8\mathrm{m} \times 10\mathrm{m}$ floorplan discretized at $0.1\mathrm{m}$ and $10^{\circ}$ yields $80 \times 100 \times 36$ camera pose candidates. To exhaustively evaluate these candidates, they further require preprocessing of floorplans, \eg, converting floorplans into ray-based representations. In contrast, our method directly operates on raw rasterized floorplans without such constraints.

\paragraphdot{Quantitative Evaluation}
As shown in Table~\ref{table:structured3d}, our method achieves state-of-the-art single-view localization accuracy despite operating under a significantly more general setting. The proposed method outperforms the strongest baseline, UnLoc~\cite{UnLoc}, by 37.2\% (51.6 vs. 37.6) on the combined angular-positional recall, demonstrating the versatility of our method alongside in-the-wild scenarios.

\begin{table}[!t]
    \centering

    \caption{\textbf{Single-view indoor localization on Structured3D~\cite{Structured3D}.} Recent localization methods rely on pre-defined discretized pose spaces and require preprocessing of floorplans, whereas our method directly operates on raw floorplans while achieving state-of-the-art results.}
    \label{table:structured3d}
    \vspace{0.5mm}
    
    \renewcommand{\arraystretch}{1.1}
    \setlength{\tabcolsep}{6pt}
    \resizebox{\textwidth}{!}{
        \begin{tabular}{l c c ccc c c}
        
        \noalign{\hrule height 0.85pt} 
        & Discretized & Floorplan & \multicolumn{3}{c}{Positional Recall\,$\uparrow$} & \!\! & \!\!Ang.\,\&\,Pos.\,$\uparrow$\!\!
        \\ 
        
        \cline{4-6}
        \cline{8-8}
        Method & Pose Space & Preprocessing & \,@\,0.1m\, & \,@\,0.5m\, & \,@\,1m\, & \!\! & \,@\,(30$^{\circ}$,\,1m)\,
        \\
        
        \noalign{\hrule height 0.7pt}
        LASER~\cite{min2022laser} & $\checkmark$ & $\checkmark$ & 0.7 & 6.4 & 10.4 & \!\! & 8.7
        \\ 
        F$^{3}$Loc~\cite{chen2024f3loc} & $\checkmark$ & $\checkmark$ & 1.5 & 14.6 & 22.4 & \!\! & 21.3
        \\ 
        UnLoc~\cite{UnLoc} & $\checkmark$ & $\checkmark$ & \textbf{5.3} & 33.9 & 38.8 & \!\! & 37.6
        \\ 
        \cellcolor{gray!7.5}\textbf{Ours} & \cellcolor{gray!7.5}\ding{55} & \cellcolor{gray!7.5}\ding{55} & \cellcolor{gray!7.5}3.5 & \cellcolor{gray!7.5}\textbf{37.5} & \cellcolor{gray!7.5}\textbf{53.8} & \cellcolor{gray!7.5}\!\! & \cellcolor{gray!7.5}\textbf{51.6}
        \\ 
        
        \noalign{\hrule height 0.85pt}
    \end{tabular}}
\end{table}

\begin{figure}[t!]
  \centering
  \includegraphics[width=0.95\linewidth]{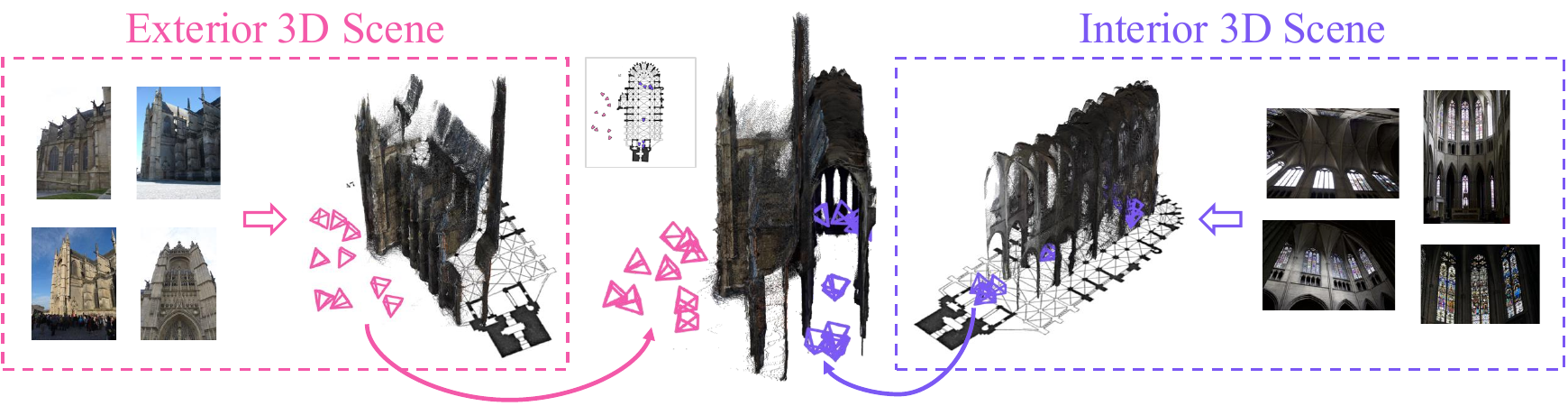}
  \caption{\textbf{Alignment of interior and exterior 3D scenes.} Using the floorplan as a shared geometric bridge, our approach enables independent alignment of interior and exterior reconstructions into a unified global coordinate system. This is achieved despite minimal visual overlap and large viewpoint differences between indoor and outdoor scenes.}
  \label{fig:interior_exterior_3d_alignment}
\end{figure}

\subsection{Downstream Applications}
\label{sec:applications}

A key advantage of floorplan-based alignment is that the floorplan provides a shared geometric reference for image collections that do not directly overlap in appearance or viewpoint. This enables separate reconstructions to be registered into a common coordinate system even when direct 3D alignment is not possible. We demonstrate this capability in the challenging setting of interior and exterior image collections of the same building, where minimal visual overlap and large viewpoint differences typically prevent coherent joint reconstruction (Figure~\ref{fig:interior_exterior_3d_alignment}). Our method also supports the alignment of disjoint image collections without any overlapping regions (Figure~\ref{fig:supp_disjoint_3d_alignment}).

%% file: sec/5_conclusion.tex
\section{Conclusion}
\label{sec:conclusion}

In this work, we introduced \methodName{}, an approach that revisits floorplan localization through the lens of contemporary visual foundation models. By leveraging the geometric priors of pretrained 3D reconstruction models together with the representational capabilities of 2D visual foundation models, our approach enables reframing floorplan localization as a reconstruction and alignment problem grounded in scene geometry. While our method demonstrates encouraging performance, its reliance on 3D foundation models presents an inherent limitation: inaccuracies in the reconstructed geometry can propagate to the density map, thereby impacting alignment quality. Future work can explore improving robustness to imperfect reconstructions and jointly reasoning about scene geometry and floorplan alignment for more reliable floorplan localization in challenging environments.

%% file: sec/supp_0_intro.tex
\section*{Appendix}

We refer readers to the accompanying \texttt{viewer.html} for 360$^{\circ}$ view comparisons of floorplan-aligned 3D reconstructions (Sec.~\ref{sec:supp_html_viewer}). In this document, we provide more dataset details (Sec.~\ref{sec:supp_dataset_statistics}), implementation details (Sec.~\ref{sec:supp_implementation_details}), qualitative results (Sec.~\ref{sec:supp_qualitative_results}), quantitative evaluations (Sec.~\ref{sec:supp_ablation_study}), analyses (Sec.~\ref{sec:supp_analyses}), limitations (Sec.~\ref{sec:supp_limitations}), applications (Sec.~\ref{sec:supp_applications}), and broader impacts (Sec.~\ref{sec:supp_broader_impacts}), which are not included in the main paper due to space constraints.

\setcounter{figure}{0}
\setcounter{table}{0}
\renewcommand{\thefigure}{A.\arabic{figure}}
\renewcommand{\thetable}{A.\arabic{table}}

\section{\texorpdfstring{360$^{\boldsymbol{\circ}}$ View Comparisons of Floorplan Alignment}{360-degree View Comparisons of Floorplan Alignment}}
\label{sec:supp_html_viewer}

The accompanying HTML file shows 360$^{\circ}$ view renderings of floorplan-aligned 3D reconstructions, allowing visual comparison of alignment results before and after fine-tuning  DINOv3~\cite{DINOv3}.

\begin{figure}[h!]
  \centering
  \includegraphics[width=\linewidth]{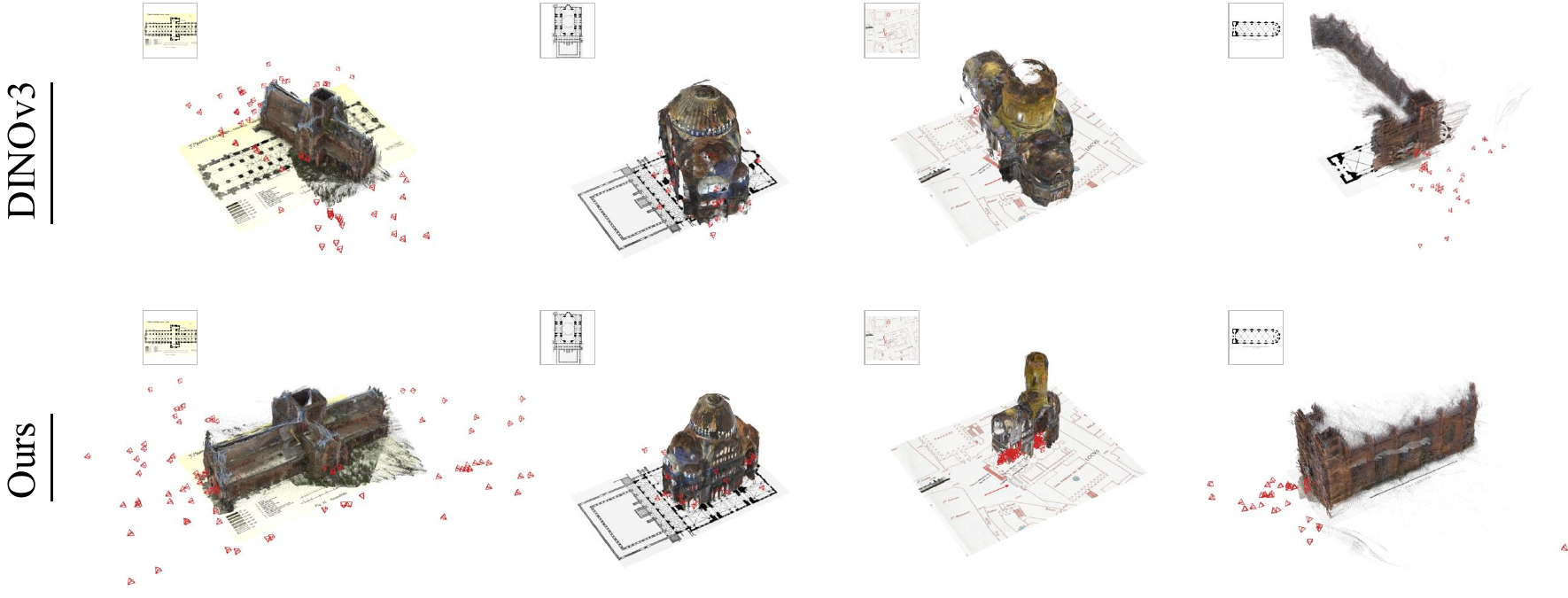}
  \vspace{-4mm}
  \caption{\textbf{Preview.} We compare floorplan alignment results using correspondences estimated by pretrained DINOv3~\cite{DINOv3} and our fine-tuned model.}
  \label{fig:supp_html_teaser}
\end{figure}

%% file: sec/supp_1_dataset_statistics.tex
\setcounter{figure}{0}
\setcounter{table}{0}
\renewcommand{\thefigure}{B.\arabic{figure}}
\renewcommand{\thetable}{B.\arabic{table}}

\section{In-the-Wild Dataset Details}
\label{sec:supp_dataset_statistics}

Table~\ref{tab:supp_dataset_statistics} shows statistics for the C3 dataset~\cite{C3Po} used in our experiments. The original dataset contains samples with severe annotation errors due to noise in Structure-from-Motion reconstructions and their registration to floorplans. To establish a reliable benchmark, we construct a clean subset by filtering such samples, retaining $463$ of $479$ training scenes and $111$ of $118$ test scenes. Furthermore, we distinguish view types of the scenes by annotating each image as \texttt{interior} or \texttt{exterior}, resulting in $738$ unique scenarios. Each scenario is defined by a triplet $\langle \mathrm{s, v, f}\rangle$, where $s$ denotes the scene, $v \in \{\mathrm{interior, exterior}\}$ indicates the scene view type, and $f$ denotes the floorplan.

\begin{figure}[t!]
  \centering
  \includegraphics[width=\linewidth]{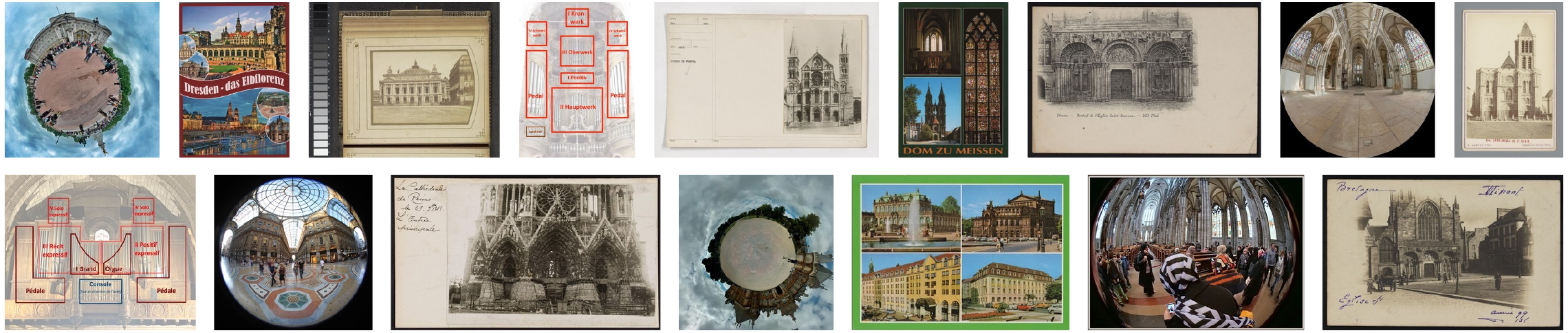}
  \vspace{-3mm}
  \caption{\textbf{Examples of filtered samples.} 
  We show samples removed from the original C3 dataset~\cite{C3Po} by our filtering pipeline. For example, images with severe optical distortions (\eg, fisheye lenses) and non-photographic content (\eg, blueprints, illustrations) are excluded.}
  \label{fig:supp_bad_samples}
\end{figure}

\paragraphdot{Filtering Strategy}
We curate the clean subset using a multi-stage pipeline designed to remove noise with varying degrees of detectability. We first apply a set of geometric and heuristic filters to eliminate duplicates, images with extreme aspect ratios, and pairs with too few correspondences for reliable matching or excessive out-of-boundary correspondences. These algorithmic filters remove obvious low-quality samples. We then use Qwen2.5-VL~\cite{bai2025qwen25vltechnicalreport} to identify failure cases that are harder to capture with simple rules, such as severe optical distortions (\eg, fisheye images) and non-photographic inputs (\eg, blueprints). Finally, we conduct two rounds of visual inspection at both image and scene levels to remove remaining problematic samples. Figure~\ref{fig:supp_bad_samples} shows examples of filtered samples.

\paragraphdot{Interior/Exterior Labeling}
We assign each image an \texttt{interior} or \texttt{exterior} label through a multi-stage pipeline. First, we employ Qwen2.5-VL~\cite{bai2025qwen25vltechnicalreport} to classify each image as \texttt{interior}, \texttt{exterior}, or \texttt{unsure} based on camera location and viewpoint. For \texttt{unsure} cases, tentative labels are assigned using CLIP~\cite{radford2021learning} as a zero-shot classifier. Predicted labels are then overridden when structural keywords (\eg, ``facade'', ``nave'') appear in the file path. Finally, we conduct three rounds of visual inspection to correct any remaining misclassifications.

\begin{table*}[t!]
    \centering
    \caption{\textbf{Dataset statistics of the clean subset from C3~\cite{C3Po}.} Scene-level and image-level statistics are reported, grouped by interior and exterior view types. In the Subtype column, \texttt{Scene} denotes the number of unique scenes. \texttt{Scene-Floorplan} counts scene-floorplan pairs, as a scene may be associated with multiple floorplans. \texttt{Scene-\{Interior, Exterior\}-Floorplan} further distinguishes view types within each pair, since images from the same scene can be categorized as either \texttt{Interior} or \texttt{Exterior}.}
    \label{tab:supp_dataset_statistics}
    \vspace{0.5mm}
    \renewcommand{\arraystretch}{1.05}
    \setlength{\tabcolsep}{8pt}
    \resizebox{\linewidth}{!}{
        \begin{tabular}{l l l r}
        \noalign{\hrule height 0.85pt}
        Split & Level\;\;\;\;\; & Subtype & Count \\
        \noalign{\hrule height 0.7pt}
        \multirow{7}{*}{Train\;\;} 
        & \multirow{5}{*}{Scene} & \texttt{Scene} & 463 \\
        & & \hspace{0.25em} $\llcorner$ \texttt{Scene-Floorplan} & 498 \\
        & & \hspace{1.25em} $\llcorner$ \texttt{Scene-\{Interior, Exterior\}-Floorplan}\;\; & 598 \\
        & & \hspace{2.25em} $\llcorner$ \texttt{Scene-Interior-Floorplan} & 186 \\
        & & \hspace{2.25em} $\llcorner$ \texttt{Scene-Exterior-Floorplan} & 412 \\
        \cline{2-4}
        & \multirow{2}{*}{Image} & \texttt{Interior} & 14,781 \\
        & & \texttt{Exterior} & 38,096 \\
        \noalign{\hrule height 0.7pt}
        \multirow{7}{*}{Test\;\;} 
        & \multirow{5}{*}{Scene} & \texttt{Scene} & 111 \\
        & & \hspace{0.25em} $\llcorner$ \texttt{Scene-Floorplan} & 120 \\
        & & \hspace{1.25em} $\llcorner$ \texttt{Scene-\{Interior, Exterior\}-Floorplan}\;\; & 140 \\
        & & \hspace{2.25em} $\llcorner$ \texttt{Scene-Interior-Floorplan} & 38 \\
        & & \hspace{2.25em} $\llcorner$ \texttt{Scene-Exterior-Floorplan} & 102 \\
        \cline{2-4}
        & \multirow{2}{*}{Image} & \texttt{Interior} & 5,659 \\
        & & \texttt{Exterior} & 6,555 \\
        \noalign{\hrule height 0.85pt}
        \end{tabular}
    }
\end{table*}

%% file: sec/supp_2_implementation_details.tex
\setcounter{figure}{0}
\setcounter{table}{0}
\renewcommand{\thefigure}{C.\arabic{figure}}
\renewcommand{\thetable}{C.\arabic{table}}

\section{Additional Implementation Details}
\label{sec:supp_implementation_details}

This section provides additional implementation details regarding our cross-modal correspondence estimation for floorplan alignment, extending Section~\ref{sec:experimental_setup} of the main paper. The following subsections describe our setup for in-the-wild evaluation. We detail the architecture of the correspondence estimation network (Sec.~\ref{sec:model_architecture}), followed by data augmentation strategies (Sec.~\ref{sec:supp_data_augmentation}) and training details (Sec.~\ref{sec:supp_training_details}). We then specify the inference settings used to predict a 2D similarity transform (Sec.~\ref{sec:supp_inference_settings}). The setup for synthetic indoor evaluation is provided in Section~\ref{sec:supp_structured3d_setup}.

\subsection{Model Architecture}
\label{sec:model_architecture}

\paragraphdot{Correspondence Estimation Network}
We use DINOv3 ViT-B/16~\cite{DINOv3} as the shared encoder $\mathcal{E}$. To adapt the encoder to our task while preserving its pretrained visual representations, we freeze the backbone weights and inject trainable Low-Rank Adaptation (LoRA)~\cite{LoRA} layers into the attention and MLP modules of each transformer block, with rank $r = 16$, scaling factor $\alpha = 16$, and dropout rate $0.1$. Given a density map $D \in \mathbb{R}^{H \times W \times 1}$ and a reference floorplan $F \in \mathbb{R}^{H \times W \times 3}$, the encoder produces $C$-dimensional feature maps with spatial resolution $H/16 \times W/16$, where $C=768$. The input resolution $H \times W$ is fixed to $512 \times 512$ at inference time, but we vary it during training.

\subsection{Data Augmentation}
\label{sec:supp_data_augmentation}

\begin{figure}[t!]
  \centering
  \includegraphics[width=\linewidth]{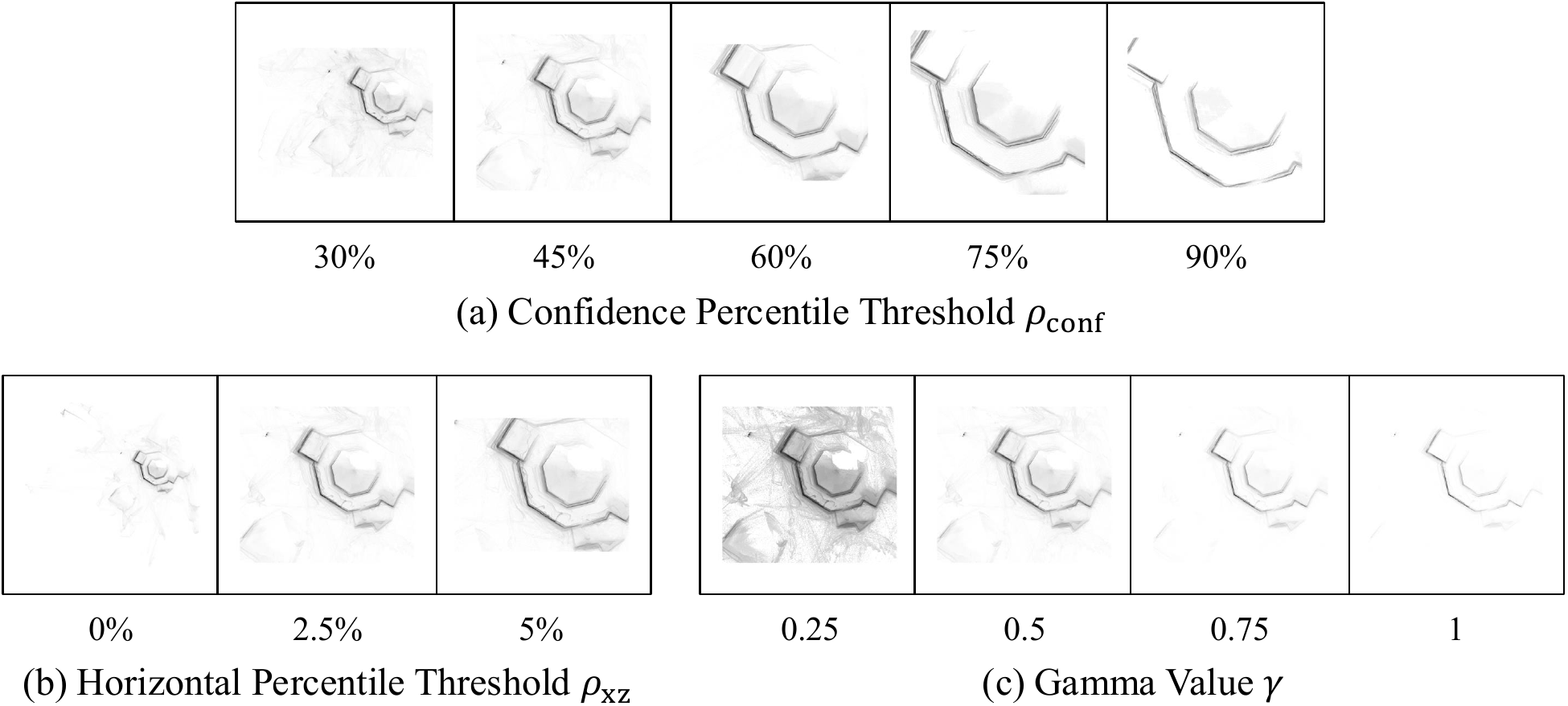}
  \vspace{-3mm}
  \caption{\textbf{Extracting density maps across varying hyperparameters.} We visualize density maps obtained by varying the confidence percentile threshold $\rho_{\mathrm{conf}}$, horizontal percentile threshold $\rho_{xz}$, and gamma value $\gamma$, with others fixed at their default values ($\rho_{\mathrm{conf}} = 45\%$, $\rho_{xz} = 2.5\%$, $\gamma = 0.5$).}
  \label{fig:supp_data_augmentation}
  \vspace{-1.5mm}
\end{figure}

\paragraphdot{Data Augmentation for Extracting Density Maps}
To simulate the noise of in-the-wild 3D reconstructions, we derive density maps under various configurations. Specifically, we use two reconstruction models ($\pi^{3}$~\cite{Pi3} and VGGT~\cite{VGGT})
\footnote{We use the model weights from~\cite{li2026longtailinternetphotoreconstruction}, trained for robust 3D reconstruction from noisy Internet photo collections.} 
with two image preprocessing modes (cropping and padding) that produce $518 \times 518$ input images. For each 3D scene reconstruction, we randomly sample between $50$ and $150$ images, while scenes with fewer images are processed using all available ones. As shown in Figure~\ref{fig:supp_data_augmentation}, we also randomize these hyperparameters: reconstruction confidence thresholds $\rho_{\mathrm{conf}} \in \{30\%, 45\%, 60\%, 75\%, 90\%\}$, horizontal percentile thresholds $\rho_{xz} \in \{0\%, 2.5\%, 5\% \}$, and gamma values $\gamma \in \{0.25, 0.5, 0.75, 1\}$. Combined with the height slicing percentiles $\left[\rho_{y}^{\mathrm{min}}, \rho_{y}^{\mathrm{max}}\right] = \left[20\%, 95\%\right]$, these configurations yield $\sim$$140K$ density maps. Furthermore, $\sim$$140K$ density maps from single-view reconstructions are included to effectively handle sparse-view scenarios.

\paragraphdot{Data Augmentation on Density Maps and Floorplans}
During training, we apply on-the-fly augmentations to density maps and floorplans. For geometric augmentations, we use random rotations sampled from $\{0^{\circ}, 90^{\circ}, 180^{\circ}, 270^{\circ}\}$ and random cropping with a scale factor of $[0.7, 1.0]$ from a randomly selected top-left corner. For photometric augmentations, 
we apply color jittering: floorplans are perturbed with brightness $0.5$, contrast $0.5$, saturation $0.5$, and hue $0.1$, whereas density maps are adjusted with brightness $0.2$ and contrast $0.2$.

\paragraphdot{Mixed-Resolution Training Strategy} 
DINOv3~\cite{DINOv3} utilizes Rotary Positional Embeddings (RoPE)~\cite{su2024roformer}, allowing the model to process variable input resolutions without architectural modifications. We leverage this property and dynamically vary input resolutions during training to improve correspondence matching across diverse floorplans with varying target region sizes. All images within each batch share the same spatial resolution, which is randomly selected per batch from $\{ 512\times512,\, 768\times768,\, 1024\times1024,\, 1280\times1280 \}$.

\subsection{Training Details}
\label{sec:supp_training_details}

\paragraphdot{Loss Coefficients and Curriculum Learning}
At each training iteration, $Q=1024$ randomly sampled correspondence pairs are used for loss calculation. The loss coefficients are set to $\lambda_{\mathrm{feat}}=1, \lambda_{\mathrm{regr}}=50, \lambda_{\mathrm{topo}}=10,$ and $\lambda_{\mathrm{geo}}=10$. We set the temperature parameter to $\tau=0.07$ for the feature matching loss. To stabilize training, we employ a curriculum learning strategy. During the first $10\%$ of training iterations, the network is trained only using the feature matching loss $\mathcal{L}_{\mathrm{feat}}$. From $10\%$ to $20\%$, the coordinate regression loss $\mathcal{L}_{\mathrm{regr}}$ is introduced. For the remaining iterations, we also incorporate the structural regularization terms $\mathcal{L}_{\mathrm{topo}}$ and $\mathcal{L}_{\mathrm{geo}}$.

\paragraphdot{Optimization}
The network is optimized using the AdamW optimizer~\cite{AdamW} with a learning rate of $10^{-4}$ and a batch size of $4$. Gradient clipping with a maximum norm of $1.0$ is applied for stable optimization. We train the network for $1$ epoch on a total of $\sim$$280K$ augmented samples, which takes $\sim$$2$ days on a single NVIDIA RTX A6000 GPU.

\subsection{Inference Settings}
\label{sec:supp_inference_settings}

To handle large-scale 3D scenes under GPU memory constraints, scenes with $N > 150$ images are partitioned into $\lceil N / 150 \rceil$ chunks, whereas smaller scenes are processed entirely. For 3D scene reconstruction, input images are preprocessed with padding. For density map extraction, we set $\rho_{\mathrm{conf}} = 45\%$, $\rho_{xz} = 2.5\%$, $\gamma = 0.5$, and $\left[\rho_{y}^{\mathrm{min}}, \rho_{y}^{\mathrm{max}}\right] = \left[20\%, 95\%\right]$. To estimate a 2D similarity transform $\mathbf{M}$ between a density map $D$ and a floorplan $F$, we sample query points from $D$ with probability proportional to the density values, and identify reliable correspondences by retaining the top 50\% confidence points and applying mutual nearest neighbor matching. The remaining correspondences are used to estimate $\mathbf{M}$ via RANSAC~\cite{fischler1981random}. We then apply $\mathbf{M}$ to the reconstructed 3D points $\mathcal{P}^{\mathrm{g}}_{*}$ and their associated camera poses, yielding a floorplan-aligned 3D scene.

\subsection{Structured3D Setup}
\label{sec:supp_structured3d_setup}

For Structured3D~\cite{Structured3D}, we adopt the same model hyperparameters, loss functions, and optimization settings as used for in-the-wild setup. We replace the training data with single-view samples from the Structured3D training set, where each density map is simply extracted using our inference settings (Sec.~\ref{sec:supp_inference_settings}) without augmentations. As this slows convergence, we train the network for $10$ epochs.

%% file: sec/supp_3_qualitative_results.tex
\setcounter{figure}{0}
\setcounter{table}{0}
\renewcommand\thefigure{D.\arabic{figure}}
\renewcommand\thetable{D.\arabic{table}}

\begin{figure}[t!]
  \centering
  \includegraphics[width=\linewidth]{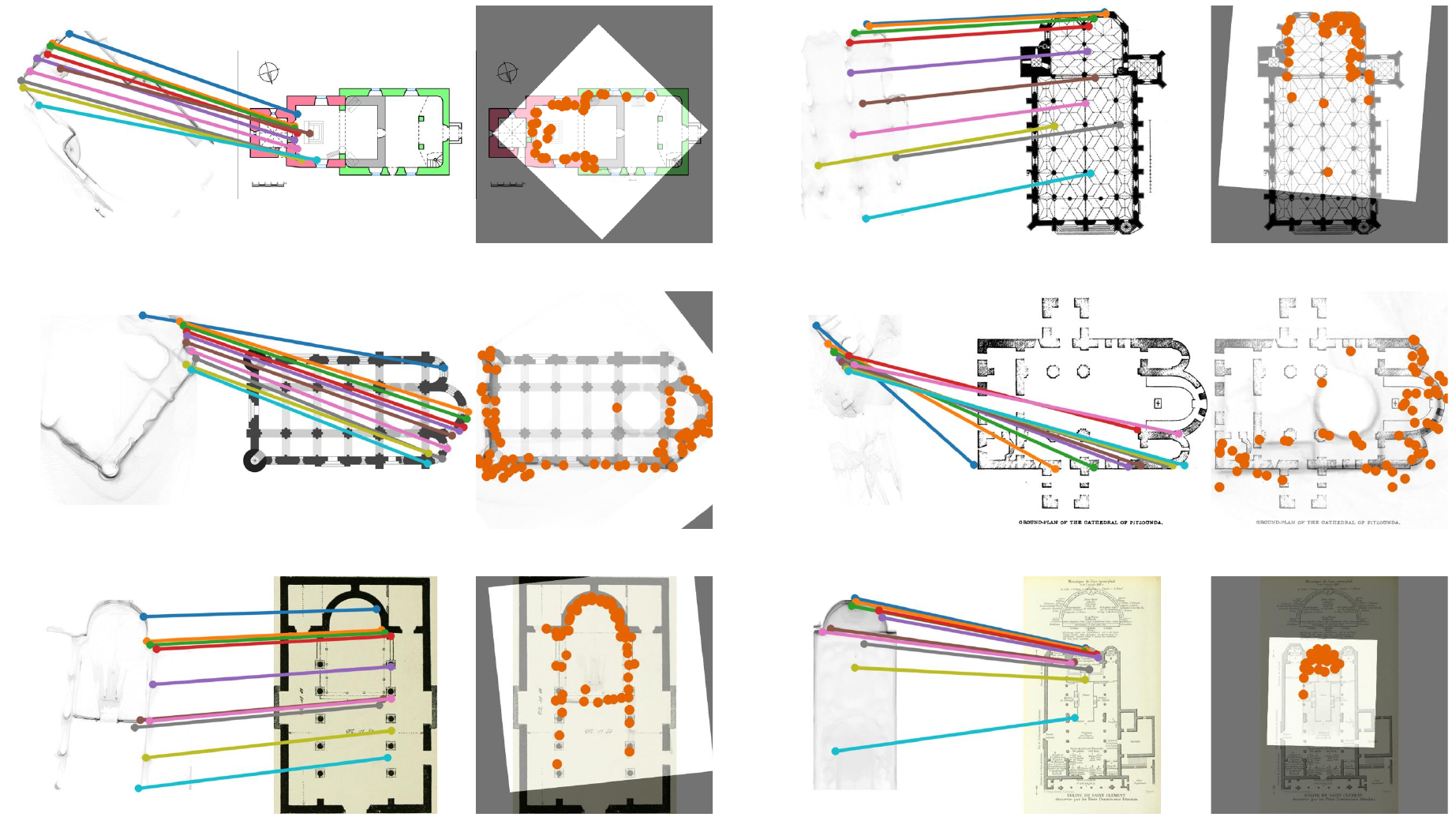}
  \vspace{-3mm}
  \caption{\textbf{Floorplan alignment via correspondence estimation.} We predict a 2D similarity transform $\mathbf{M}$ for floorplan alignment by estimating correspondences between a density map and a reference floorplan. Here, we visualize 10 randomly sampled correspondences predicted by our model. Reliable correspondences used to compute the similarity transform are overlaid (in \textcolor{corrcol}{orange}) on the aligned density map.}
  \label{fig:supp_correspondence_pair_warping}
  \vspace{-1mm}
\end{figure}

\section{Additional Qualitative Results}
\label{sec:supp_qualitative_results}

\begin{figure}[t!]
  \centering
  \includegraphics[width=\linewidth]{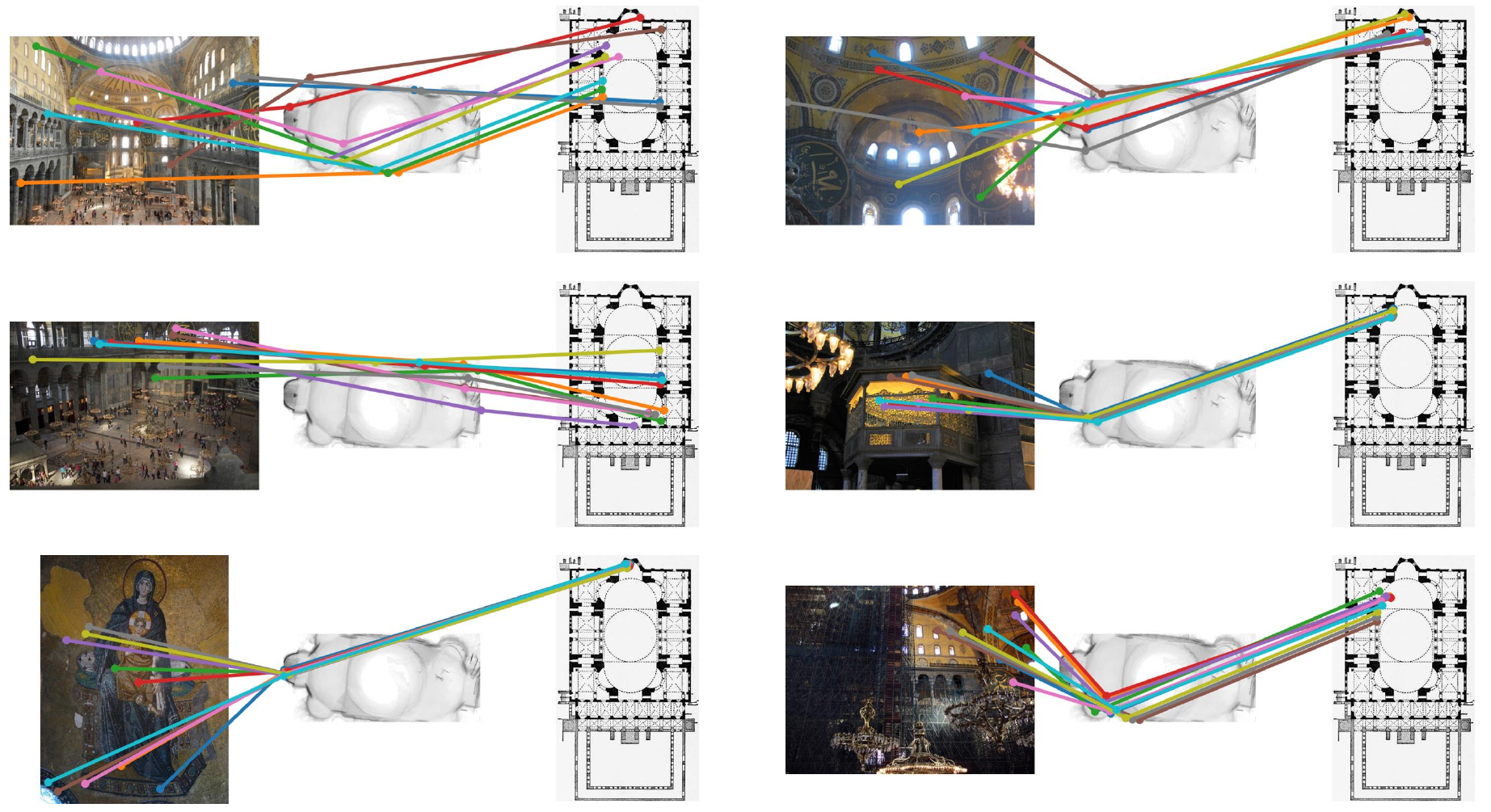}
  \vspace{-2mm}
  \caption{\textbf{Photo-to-Floorplan correspondences via density map.}
  We localize pixel-level observations by aligning a density map to a reference floorplan. Here, we visualize 10 randomly sampled correspondences. The matches from photo to density map are obtained via 3D reconstruction, and those from density map to floorplan are predicted via cross-modal correspondence estimation.}
  \label{fig:supp_correspondence_triple}
  \vspace{2mm}
\end{figure}

This section provides more qualitative results that illustrate how our method performs floorplan localization. As shown in Figure~\ref{fig:supp_correspondence_pair_warping}, we predict correspondences between a density map and a reference floorplan to estimate a 2D similarity transform for floorplan alignment. The results demonstrate that our model can handle floorplans with varying target region sizes. Figure~\ref{fig:supp_correspondence_triple} shows how we estimate pixel-level correspondences between a photo and a floorplan through a density map. After we align the density map to the floorplan, we can localize pixel-level observations as we know the mapping from photo to density map (via 3D reconstruction) and from density map to floorplan (via correspondence estimation). Since the density map captures the global context of the scene, our method can effectively localize minimal-context photos as shown in the last row on the left.

\begin{figure}[t!]
  \centering
  \includegraphics[width=\linewidth]{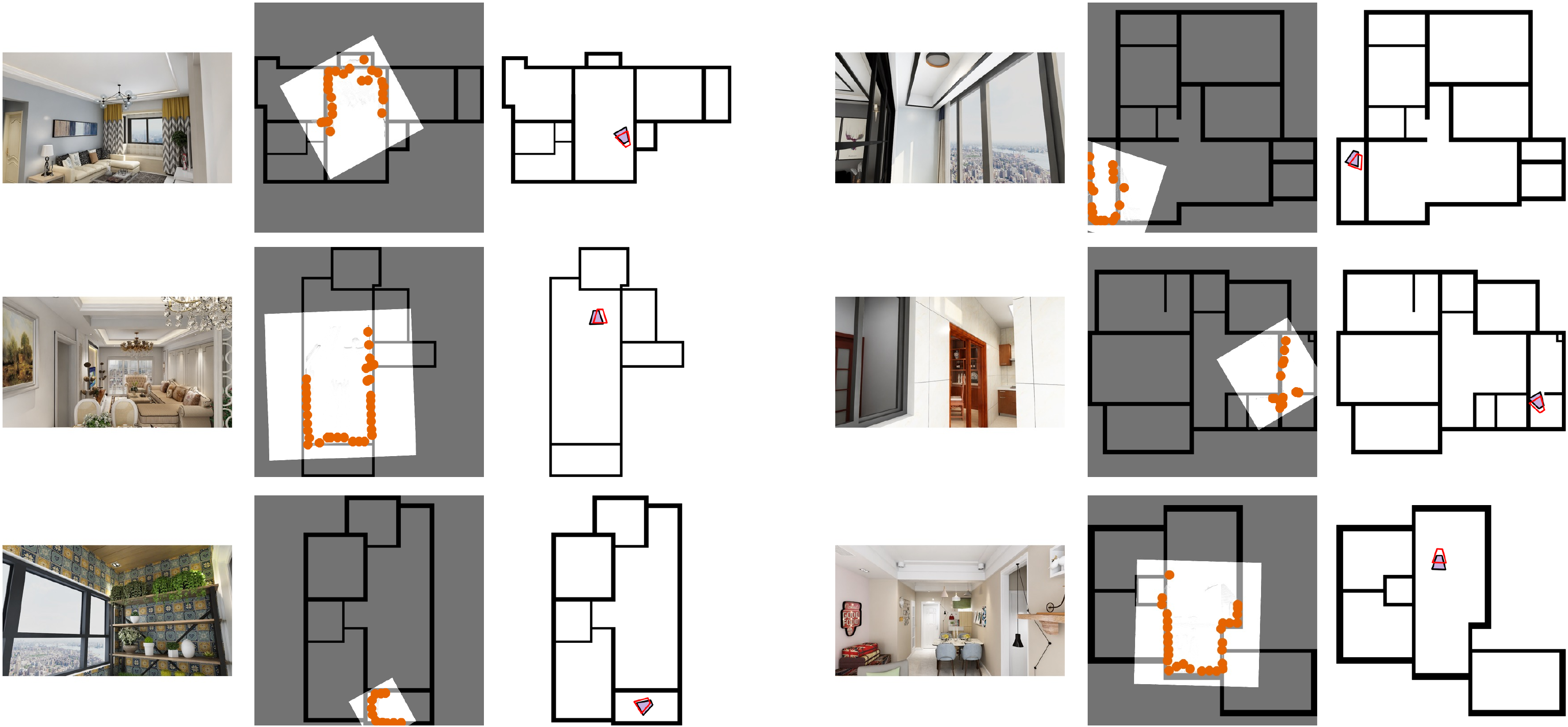}
  \vspace{-1mm}
  \caption{\textbf{Qualitative results on Structured3D~\cite{Structured3D}.} For each sample, we show the input photo, floorplan-aligned density map (with reliable correspondences overlaid in \textcolor{corrcol}{orange}), and floorplan localization result. Cameras are illustrated as {\protect\camicon{red}{white}} for \textcolor{red}{GT} and {\protect\camicon{black}{ourslightcol}} for \textcolor{ourscol}{Ours}. Despite the limited visual context from a single image, our method successfully recovers camera poses by aligning the extracted density map to the reference floorplan.}
  \label{fig:supp_structured_3d}
  \vspace{0.5mm}
\end{figure}

We further provide qualitative results on Structured3D~\cite{Structured3D} in Figure~\ref{fig:supp_structured_3d}, demonstrating that our method extends to single-view indoor localization despite operating on rasterized floorplans without a discretized pose space or floorplan preprocessing.

\clearpage
\begin{figure}[t!]
  \centering
  \includegraphics[width=\linewidth]{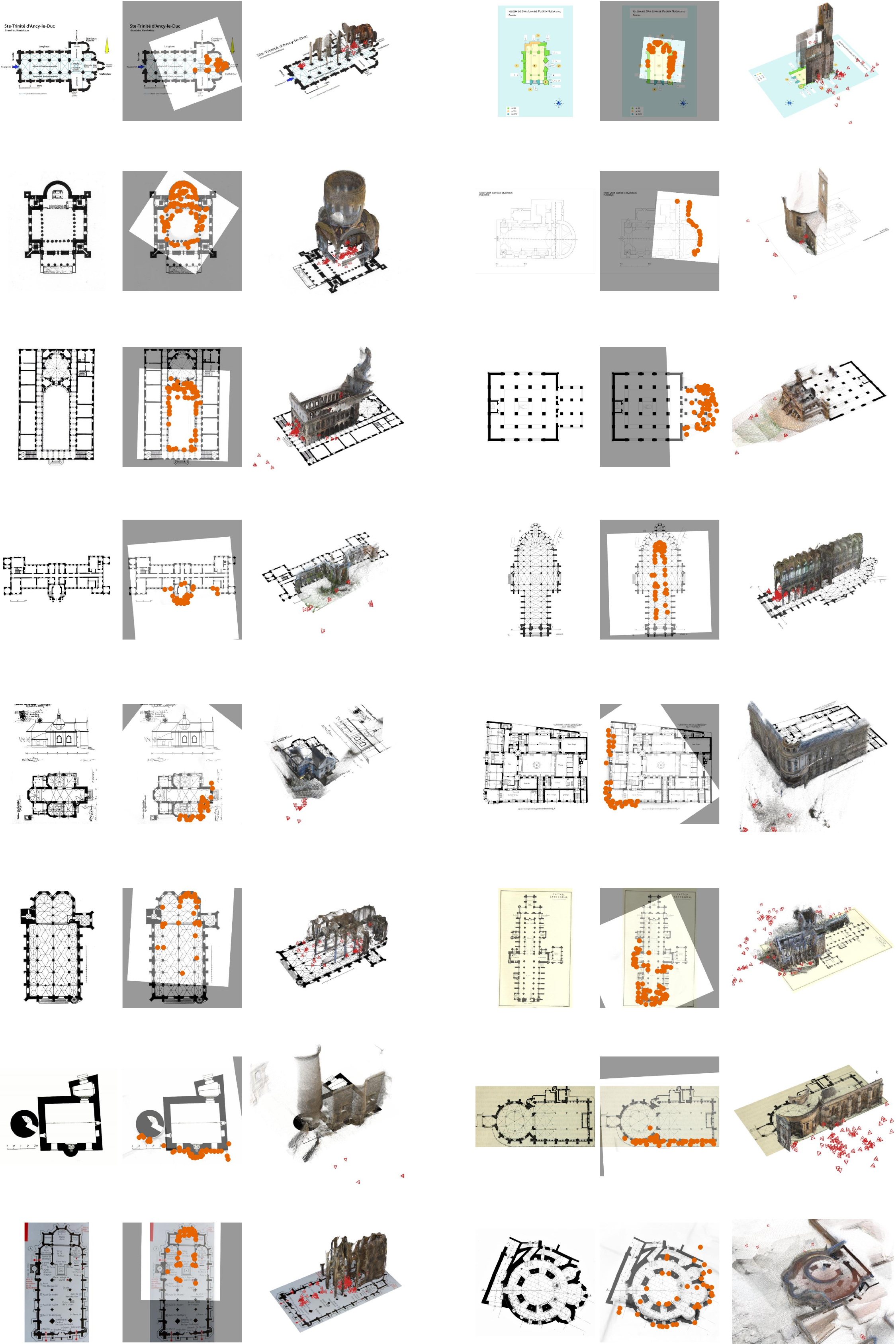}
  \vspace{2mm}
  \caption{\textbf{Floorplan-aligned 3D scene reconstructions.} For each sample, we show the input floorplan, aligned density map (with reliable correspondences overlaid in \textcolor{corrcol}{orange}), and aligned scene.}
  \label{fig:supp_floorplan_aligned_3d_scenes}
\end{figure}
\clearpage

%% file: sec/supp_4_ablation_study.tex
\setcounter{figure}{0}
\setcounter{table}{0}
\renewcommand\thefigure{E.\arabic{figure}}
\renewcommand\thetable{E.\arabic{table}}

\section{Additional Quantitative Evaluations on In-the-Wild Data}
\label{sec:supp_ablation_study}

In this section, we provide additional evaluations to validate the contribution of our design choices (Sec.~\ref{sec:supp_design_choices}), including the ablation study on learning objectives (Table~\ref{table:loss_ablation}), correspondence filtering strategies (Table~\ref{table:effect_of_correspondence_filtering}), LoRA configurations (Table~\ref{table:supp_lora_hyperparameters}), mixed-resolution training (Table~\ref{table:supp_mixed_resolution_training}), 3D reconstruction models (Table~\ref{table:supp_effect_of_3d_recon_models}), and density map--floorplan matching (Table~\ref{table:supp_correspondence_density_map_floorplan}). We also examine the robustness of our model against various hyperparameters (Sec.~\ref{sec:supp_robustness_to_density_map_hyperparameters}).

\begin{table*}[!t]
    \centering
    \caption{\textbf{Ablation study on loss design.}
    We evaluate each objective's contribution to cross-modal correspondence learning by progressively adding the feature matching loss $\mathcal{L}_{\mathrm{feat}}$, coordinate regression loss $\mathcal{L}_{\mathrm{regr}}$, topology preservation loss $\mathcal{L}_{\mathrm{topo}}$, and geometry consistency loss $\mathcal{L}_{\mathrm{geo}}$.} 
    \label{table:loss_ablation}
    \renewcommand{\arraystretch}{1.1}
    \setlength{\tabcolsep}{2.5pt}
    \resizebox{\textwidth}{!}{
        \begin{tabular}{cccc c cc c cc c c c cc c c}
        
        \noalign{\hrule height 0.85pt}
        & & & &
        \!\!\! & \multicolumn{2}{c}{Ang.\,$\uparrow$} & \!\! & \multicolumn{2}{c}{Pos.\,$\uparrow$} & \!\! & \!\!\!Ang.\,\&\,Pos.\,$\uparrow$\!\!\! & \! & \multicolumn{2}{c}{PCK\,$\uparrow$} & \!\!\! & 
        \\
        
        \cline{6-7} \cline{9-10} \cline{12-12} \cline{14-15}
        $\mathcal{L}_{\mathrm{feat}}$ & $\mathcal{L}_{\mathrm{regr}}$ & $\mathcal{L}_{\mathrm{topo}}$ & $\mathcal{L}_{\mathrm{geo}}$ & \!\!\! & \,@\,$5^\circ$ & \,@\,$30^\circ$ & \!\!\! & \,@\,$5\%$ & \,@\,$20\%$ & \!\!\! & \!\!\,@\,($30^\circ,20\%$)\!\! & \!\!\! & \,@\,1\%\, & \,@\,30\%\, & \!\!\! & \,RMSE\,$\downarrow$\,
        \\
        
        \noalign{\hrule height 0.7pt}
        
        & & & & \!\!\! & 14.25 & 35.33 & \!\!\! & 6.75 & 26.67 & \!\!\! & 18.28 & \!\!\! & 0.73 & 47.01 & \!\!\! & 0.2624
        \\
        
        \checkmark & & & & \!\!\! & 59.23 & 80.08 & \!\!\! & 48.71 & 72.42 & \!\!\! & 69.10 & \!\!\! & 20.01 & 86.85 & \!\!\! & 0.0900
        \\
        \checkmark & \checkmark & & & \!\!\! & 60.50 & 80.17 & \!\!\! & 48.74 & 73.78 & \!\!\! & 70.60 & \!\!\! & \textbf{21.12} & 86.39 & \!\!\! & 0.0938
        \\
        \checkmark & \checkmark & \checkmark & & \!\!\! & 64.18 & 80.84 & \!\!\! & 49.63 & 75.24 & \!\!\! & 70.54 & \!\!\! & 20.00 & 89.40 & \!\!\! & 0.0818
        \\
        
        \checkmark & \checkmark & \checkmark & \checkmark & \!\!\! & \textbf{65.91} & \textbf{83.08} & \!\!\! & \textbf{50.87} & \textbf{76.08} & \!\!\! & \textbf{73.58} & \!\!\! & 20.42 & \textbf{90.45} & \!\!\! & \textbf{0.0776} 
        \\
        
        \noalign{\hrule height 0.85pt}
    \end{tabular}}
\end{table*}

\begin{table*}[!t]
    \centering
    \caption{\textbf{Effect of correspondence filtering.} 
    We evaluate various filtering strategies to obtain reliable correspondences before estimating the similarity transform via RANSAC. \texttt{top$_{50}$} indicates retaining the top 50\% of correspondences ranked by the confidence $w_{i}$, and \texttt{MNN} denotes mutual nearest neighbor matching.}
    \label{table:effect_of_correspondence_filtering}
    \renewcommand{\arraystretch}{1.1}
    \setlength{\tabcolsep}{3.5pt}
    \resizebox{\linewidth}{!}{
        \begin{tabular}{cc c cc c cc c c c cc c c}
        
        \noalign{\hrule height 0.85pt}
        & & \!\!\! & \multicolumn{2}{c}{Ang.\,$\uparrow$} & \!\! & \multicolumn{2}{c}{Pos.\,$\uparrow$} & \!\! & \!\!\!Ang.\,\&\,Pos.\,$\uparrow$\!\!\! & \! & \multicolumn{2}{c}{PCK\,$\uparrow$} & \!\!\! & 
        \\
        
        \cline{4-5} \cline{7-8} \cline{10-10} \cline{12-13}
        \texttt{top$_{50}$} & \texttt{MNN} & \!\!\! & \,@\,$5^\circ$ & \,@\,$30^\circ$ & \!\!\! & \,@\,$5\%$ & \,@\,$20\%$ & \!\!\! & \!\!\,@\,($30^\circ,20\%$)\!\! & \!\!\! & \,@\,1\%\, & \,@\,30\%\, & \!\!\! & \,RMSE\,$\downarrow$\,
        \\
        
        \noalign{\hrule height 0.7pt}
        
        &  & \!\!\! & 62.19 & 80.31 & \!\!\! & 47.77 & 74.47 & \!\!\! & 70.19 & \!\!\! & 17.44 & 89.75 & \!\!\! & 0.0832
        \\
        
        \checkmark & & \!\!\! & 65.38 & 81.88 & \!\!\! & \textbf{51.16} & 75.19 & \!\!\! & 71.96 & \!\!\! & 19.08 & 89.34 & \!\!\! & 0.0806
        \\

        & \checkmark & \!\!\! & 65.66 & \textbf{83.10} & \!\!\! & 48.19 & 74.60 & \!\!\! & 71.67 & \!\!\! & 17.63 & 89.62 & \!\!\! & 0.0812
        \\
        
        \checkmark & \checkmark & \!\!\! & \textbf{65.91} & 83.08 & \!\!\! & 50.87 & \textbf{76.08} & \!\!\! & \textbf{73.58} & \!\!\! & \textbf{20.42} & \textbf{90.45} & \!\!\! & \textbf{0.0776} 
        \\
        
        \noalign{\hrule height 0.85pt}
    \end{tabular}}
\end{table*}

\subsection{Design Choices}
\label{sec:supp_design_choices}

\paragraphdot{Fine-tuning Ablations} 
As summarized in Table~\ref{table:loss_ablation}, we evaluate the effectiveness of fine-tuning with the proposed objectives by progressively integrating them into our cross-modal correspondence learning scheme. The feature matching loss $\mathcal{L}_{\mathrm{feat}}$ establishes semantic alignment, which significantly enhances the pretrained DINOv3~\cite{DINOv3}. The coordinate regression loss $\mathcal{L}_{\mathrm{regr}}$ further transitions the model from estimating discrete grid indices to continuous spatial coordinates. To prevent degenerate similarity solutions where the model satisfies the regression objective through structurally inconsistent mappings, we incorporate the structural consistency regularization losses $\mathcal{L}_{\mathrm{topo}}$ and $\mathcal{L}_{\mathrm{geo}}$. These losses explicitly penalize non-rigid deformations in terms of triplet angles and pairwise distances, which should be invariant under similarity transformations. The full objective ensures the learned correspondences are both semantically aligned and structurally consistent.

\paragraphdot{Inference Ablations}
Before estimating the 2D similarity transform, we identify reliable correspondences to ensure a high-quality inlier set for RANSAC-based estimation. Specifically, we apply confidence-based pruning (\texttt{top$_{50}$}) to retain only the top 50\% of correspondences ranked by the confidence $w_{i}$, discarding uncertain matches. In addition, mutual nearest neighbor (MNN) matching enforces bidirectional consistency. As demonstrated in Table~\ref{table:effect_of_correspondence_filtering}, these filtering strategies consistently enhance alignment accuracy.

\begin{table*}[!t]
    \centering
    \caption{\textbf{Effect of LoRA configurations.} We evaluate our model with different LoRA~\cite{LoRA} rank $r$ and scaling factor $\alpha$. We set $r=\alpha=16$ in our main experiments.}
    \label{table:supp_lora_hyperparameters}
    \renewcommand{\arraystretch}{1.1}
    \setlength{\tabcolsep}{3.5pt}
    \resizebox{\linewidth}{!}{
        \begin{tabular}{cc c cc c cc c c c cc c c}
        
        \noalign{\hrule height 0.85pt}
        & & \!\!\! & \multicolumn{2}{c}{Ang.\,$\uparrow$} & \!\! & \multicolumn{2}{c}{Pos.\,$\uparrow$} & \!\! & \!\!\!Ang.\,\&\,Pos.\,$\uparrow$\!\!\! & \! & \multicolumn{2}{c}{PCK\,$\uparrow$} & \!\!\! & 
        \\
        
        \cline{4-5} \cline{7-8} \cline{10-10} \cline{12-13}
        \;\;$r$\;\; & \;\;$\alpha$\;\; & \!\!\! & \,@\,$5^\circ$ & \,@\,$30^\circ$ & \!\!\! & \,@\,$5\%$ & \,@\,$20\%$ & \!\!\! & \!\!\,@\,($30^\circ,20\%$)\!\! & \!\!\! & \,@\,1\%\, & \,@\,30\%\, & \!\!\! & \,RMSE\,$\downarrow$\,
        \\
        
        \noalign{\hrule height 0.7pt}
        
        4 & 4 & \!\!\! & 50.10 & 75.70 & \!\!\! & 39.22 & 66.66 & \!\!\! & 62.40 & \!\!\! & 13.86 & 86.49 & \!\!\! & 0.1030
        \\
        
        8 & 8 & \!\!\! & 62.74 & 81.62 & \!\!\! & 50.43 & 73.85 & \!\!\! & 71.01 & \!\!\! & 21.69 & 89.47 & \!\!\! & 0.0794
        \\
        
        16 & 16 & \!\!\! & 65.91 & \textbf{83.08} & \!\!\! & \textbf{50.87} & \textbf{76.08} & \!\!\! & \textbf{73.58} & \!\!\! & 20.42 & \textbf{90.45} & \!\!\! & \textbf{0.0776} 
        \\

        32 & 32 & \!\!\! & \textbf{65.93} & 80.02 & \!\!\! & 49.30 & 74.42 & \!\!\! & 71.00 & \!\!\! & \textbf{22.70} & 87.23 & \!\!\! & 0.0873
        \\
        \noalign{\hrule height 0.85pt}
    \end{tabular}}
\end{table*}

\begin{table*}[!t]
    \centering
    \caption{\textbf{Effect of mixed-resolution training.} We evaluate the contribution of our mixed-resolution training strategy (\texttt{MR}), which improves localization accuracy.}
    \label{table:supp_mixed_resolution_training}
    \renewcommand{\arraystretch}{1.1}
    \setlength{\tabcolsep}{4pt}
    \resizebox{\linewidth}{!}{
        \begin{tabular}{c c cc c cc c c c cc c c}
        
        \noalign{\hrule height 0.85pt}
        & \!\!\! & \multicolumn{2}{c}{Ang.\,$\uparrow$} & \!\! & \multicolumn{2}{c}{Pos.\,$\uparrow$} & \!\! & \!\!\!Ang.\,\&\,Pos.\,$\uparrow$\!\!\! & \! & \multicolumn{2}{c}{PCK\,$\uparrow$} & \!\!\! & 
        \\
        
        \cline{3-4} \cline{6-7} \cline{9-9} \cline{11-12}
        \;\;\;\texttt{MR}\;\;\; & \!\!\! & \,@\,$5^\circ$ & \,@\,$30^\circ$ & \!\!\! & \,@\,$5\%$ & \,@\,$20\%$ & \!\!\! & \!\!\,@\,($30^\circ,20\%$)\!\! & \!\!\! & \,@\,1\%\, & \,@\,30\%\, & \!\!\! & \,RMSE\,$\downarrow$\,
        \\
        
        \noalign{\hrule height 0.7pt}
        
        & \!\!\! & 62.12 & 81.98 & \!\!\! & 49.39 & 75.02 & \!\!\! & 71.88 & \!\!\! & 19.61 & 88.97 & \!\!\! & 0.0832
        \\
        
        \checkmark & \!\!\! & \textbf{65.91} & \textbf{83.08} & \!\!\! & \textbf{50.87} & \textbf{76.08} & \!\!\! & \textbf{73.58} & \!\!\! & \textbf{20.42} & \textbf{90.45} & \!\!\! & \textbf{0.0776}
        \\
        \noalign{\hrule height 0.85pt}
    \end{tabular}}
\end{table*}

\paragraphdot{LoRA Configurations} 
Table~\ref{table:supp_lora_hyperparameters} shows the effect of LoRA~\cite{LoRA} rank $r$ and scaling factor $\alpha$ used to adapt DINOv3~\cite{DINOv3} to our task. When the rank is too low (\eg, $r=4$), the adaptation capacity is limited, leading to noticeably degraded performance. Increasing the rank substantially improves the image-level and pixel-level localization accuracy. Among the evaluated settings, we observe that $r=\alpha=16$ leads to the best results.

\begin{table*}[!t]
    \centering
    \caption{\textbf{Effect of 3D reconstruction models.} We compare $\pi^{3}$~\cite{Pi3} and VGGT~\cite{VGGT} as the 3D estimator for gravity-aligned 3D scene reconstruction.}
    \label{table:supp_effect_of_3d_recon_models}
    \renewcommand{\arraystretch}{1.1}
    \setlength{\tabcolsep}{4pt}
    \resizebox{\linewidth}{!}{
        \begin{tabular}{c c cc c cc c c c cc c c}
        
        \noalign{\hrule height 0.85pt}
        3D & \!\!\! & \multicolumn{2}{c}{Ang.\,$\uparrow$} & \!\! & \multicolumn{2}{c}{Pos.\,$\uparrow$} & \!\! & \!\!\!Ang.\,\&\,Pos.\,$\uparrow$\!\!\! & \! & \multicolumn{2}{c}{PCK\,$\uparrow$} & \!\!\! & 
        \\
        
        \cline{3-4} \cline{6-7} \cline{9-9} \cline{11-12}
        Estimator & \!\!\! & \,@\,$5^\circ$ & \,@\,$30^\circ$ & \!\!\! & \,@\,$5\%$ & \,@\,$20\%$ & \!\!\! & \!\!\,@\,($30^\circ,20\%$)\!\! & \!\!\! & \,@\,1\%\, & \,@\,30\%\, & \!\!\! & \,RMSE\,$\downarrow$\,
        \\
        
        \noalign{\hrule height 0.7pt}
        
        VGGT & \!\!\! & 61.48 & 81.43 & \!\!\! & 47.25 & 75.16 & \!\!\! & 71.52 & \!\!\! & 18.27 & 88.63 & \!\!\! & 0.0873
        \\
        
        $\pi^{3}$ & \!\!\! & \textbf{65.91} & \textbf{83.08} & \!\!\! & \textbf{50.87} & \textbf{76.08} & \!\!\! & \textbf{73.58} & \!\!\! & \textbf{20.42} & \textbf{90.45} & \!\!\! & \textbf{0.0776}
        \\
        \noalign{\hrule height 0.85pt}
    \end{tabular}}
\end{table*}

\paragraphdot{Mixed-Resolution Training}
Table~\ref{table:supp_mixed_resolution_training} shows that training with varying input resolutions improves the image-level and pixel-level localization accuracy. We attribute this improvement to enhanced correspondence matching across floorplans with varying target region sizes. This strategy also serves as a form of data augmentation that helps mitigate overfitting during the training of our cross-modal correspondence estimation network.

\paragraphdot{3D Reconstruction Models}
Table~\ref{table:supp_effect_of_3d_recon_models} compares the effect of using different 3D foundation models for gravity-aligned 3D scene reconstruction. $\pi^{3}$~\cite{Pi3} consistently outperforms VGGT~\cite{VGGT} across all metrics, motivating its use as our default 3D reconstruction model.

\paragraphdot{Ablations on Density Map and Floorplan Correspondence}
The accuracy of floorplan alignment depends on the precision of correspondences between the density map and the reference floorplan. Table~\ref{table:supp_correspondence_density_map_floorplan} shows the contributions of our fine-tuning scheme and confidence-based filtering. Fine-tuning DINOv3~\cite{DINOv3} substantially improves matching accuracy by adapting features originally trained on natural images to top-down architectural images. When we consider only the top 50\% of correspondences ranked by the confidence $w_{i}$, matching accuracy is further improved. This observation motivates the use of confidence-based filtering for identifying reliable correspondences before predicting a 2D similarity transform.

\subsection{Robustness to Density Map Hyperparameters}
\label{sec:supp_robustness_to_density_map_hyperparameters}
\begin{figure}[t!]
  \centering
  \includegraphics[width=\linewidth]{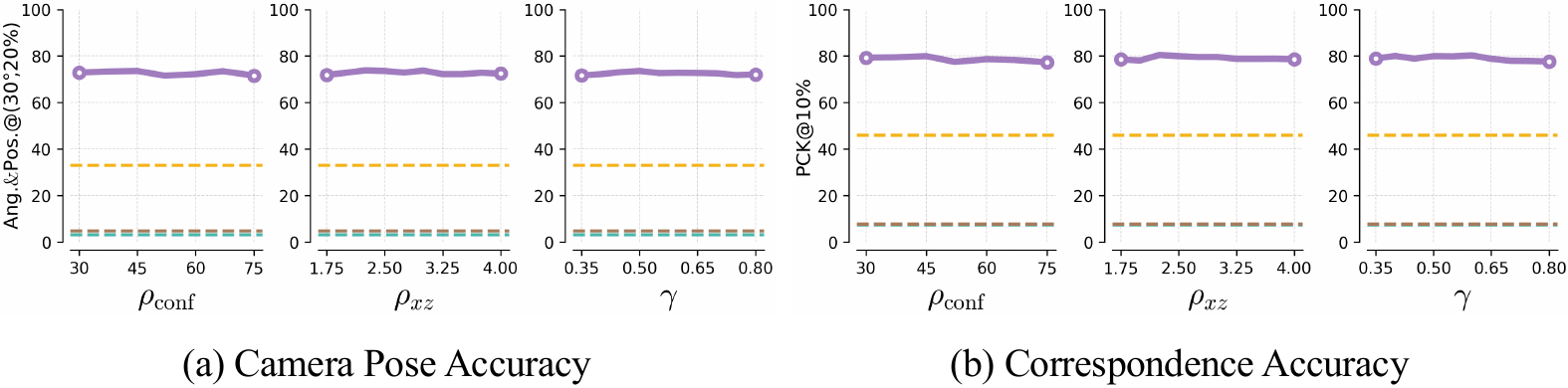}
  \vspace{-4mm}
  \caption{\textbf{Stability across hyperparameters for density map extraction.} \textcolor{ourscol}{Our model} remains robust across varying confidence percentile threshold $\rho_{\mathrm{conf}}$, horizontal percentile threshold $\rho_{xz}$, and gamma value $\gamma$. \textcolor{c3pocol}{C3Po}, \textcolor{dust3rcol}{DUSt3R}, and \textcolor{loftrcol}{LoFTR} are plotted as dashed lines for reference.}
  \label{fig:supp_different_thresholds}
\end{figure}

We evaluate the stability of our model against various hyperparameters used to derive density maps. Figure~\ref{fig:supp_different_thresholds} shows image-level and pixel-level localization accuracy when varying the confidence percentile threshold $\rho_{\mathrm{conf}}$, horizontal percentile threshold $\rho_{xz}$, and gamma value $\gamma$. Notably, our approach consistently outperforms the baselines (C3Po~\cite{C3Po}, DUSt3R~\cite{DUSt3R}, and LoFTR~\cite{LoFTR}) by a significant margin. This result indicates that the proposed method does not necessitate exhaustive hyperparameter searches to achieve state-of-the-art results, demonstrating robustness to various hyperparameters for density map extraction.

\begin{table*}[!t]
    \centering
    \caption{\textbf{Ablations on matching between density map and floorplan.} 
    We evaluate the contribution of LoRA-based fine-tuning (\texttt{FT})
    and confidence-based filtering (\texttt{top$_{50}$}) on matching between the 2D density map and reference floorplan.}
    \label{table:supp_correspondence_density_map_floorplan}
    \renewcommand{\arraystretch}{1.1}
    \setlength{\tabcolsep}{6pt}
    \resizebox{\linewidth}{!}{
        \begin{tabular}{c c cccccc c}
        
        \noalign{\hrule height 0.85pt}
        & & \multicolumn{6}{c}{PCK\,$\uparrow$} & 
        \\ 
        
        \cline{3-8}
        \texttt{FT} & \texttt{top$_{50}$} & \,@\,1\%\, & \,@\,3\%\, & \,@\,5\%\, & \,@\,10\%\, & \,@\,15\%\, & \,@\,30\%\, & \,RMSE\,$\downarrow$\,
        \\
         
        \noalign{\hrule height 0.7pt}
        
        &  & 0.79 & 4.60 & 9.13 & 20.01 & 28.58 & 53.39 & 0.2689 
        \\
        \checkmark &  & 10.54 & 48.74 & 61.67 & 69.97 & 73.81 & 87.10 & 0.1536 
        \\
        & \checkmark & 1.41 & 8.24 & 15.64 & 30.71 & 39.60 & 59.21 & 0.2690 
        \\
        \checkmark & \checkmark & \textbf{14.46} & \textbf{57.79} & \textbf{70.24} & \textbf{77.89} & \textbf{80.81} & \textbf{91.15} & \textbf{0.1290}
        \\
        \noalign{\hrule height 0.85pt}
    \end{tabular}}
\end{table*}

%% file: sec/supp_5_analysis.tex
\setcounter{figure}{0}
\setcounter{table}{0}
\renewcommand\thefigure{F.\arabic{figure}}
\renewcommand\thetable{F.\arabic{table}}

\section{More Analyses}
\label{sec:supp_analyses}

\subsection{Sparse-View 3D Reconstruction Quality}
\label{sec:supp_analyses_sparse_view_3d_recon}

In Section~\ref{sec:evaluation_in_the_wild_robustness_analysis} and Figures~\ref{fig:performance_across_varying_view_counts}-\ref{fig:visualize_sparse_recon} of the main paper, we demonstrate that our method achieves high accuracy even under sparse-view conditions. However, as discussed in Section~\ref{sec:conclusion}, the proposed approach relies on 3D foundation models~\cite{Pi3,VGGT}; geometric errors in reconstructed 3D points may propagate to density maps and thus affect floorplan alignment accuracy. To quantify the geometric degradation caused by view sparsity, we analyze how much 3D geometry reconstructed from sparse-view inputs deviates from dense-view reconstructions. In this context, we consider the dense-view reconstructions as the pseudo ground truth. The results are summarized in Table~\ref{table:supp_analysis}.

\paragraphdot{Evaluation Protocol}
To isolate the geometry of reconstructed 3D points from global camera pose drifts, we align 3D points from sparse-view and dense-view reconstructions. Specifically, we transform them into a shared camera frame using relative camera extrinsics, and further align them via Iterative Closest Point (ICP) registration. For consistent comparison, both 3D points are centered and scaled using the bounding box of the dense-view points.

\paragraphdot{Metrics}
We report \textit{Accuracy}, \textit{Completeness}, \textit{Overall}, and \textit{F-Score} using $10K$ points randomly sampled from sparse-view and dense-view reconstructions. Let $\mathcal{P}_\mathrm{pred}$ be the sampled points from the sparse-view reconstruction and $\mathcal{P}_\mathrm{gt}$ be the sampled points from the dense-view reconstruction. Accuracy is defined as the mean Euclidean distance from points in $\mathcal{P}_\mathrm{pred}$ to their nearest neighbors in $\mathcal{P}_\mathrm{gt}$. Completeness measures the mean Euclidean distance from points in $\mathcal{P}_\mathrm{gt}$ to their nearest neighbors in $\mathcal{P}_\mathrm{pred}$. Overall is defined as the average of Accuracy and Completeness, which is equivalent to the Chamfer Distance. F-Score measures the harmonic mean of precision and recall at distance thresholds, representing the concordance between points in $\mathcal{P}_\mathrm{pred}$ and $\mathcal{P}_\mathrm{gt}$.

\begin{table*}[!t]
    \centering
    \caption{\textbf{Evaluation of 3D reconstruction quality from in-the-wild sparse views.} We measure the geometric degradation of 3D points reconstructed from varying numbers of sparse input images ($1$ to $10$). Dense-view reconstructions ($\leq 150$ images) serve as the pseudo ground truth. All 3D points are estimated using $\pi^{3}$~\cite{Pi3}, then normalized and aligned via ICP.}
    \label{table:supp_analysis}
    \vspace{-0.75mm}
    \renewcommand{\arraystretch}{1.1}
    \setlength{\tabcolsep}{4.5pt}
    \resizebox{\linewidth}{!}{
        \begin{tabular}{c cccccc c}
        
        \noalign{\hrule height 0.85pt}
        & & & & \multicolumn{4}{c}{F-Score\,$\uparrow$}  
        \\ 
        
        \cline{5-8}
        \;\# Imgs\;  & \,Accuracy\,$\downarrow$\, & \,Completeness\,$\downarrow$\, & \,Overall\,$\downarrow$\, & \,@\,1\%\, & \,@\,3\%\, & \,@\,5\%\, & \,@\,10\%\,
        \\
         
        \noalign{\hrule height 0.7pt}
        
        $= 1$ & 1.1470 & 0.9222 & 1.0346 & 14.81 & 32.41 & 38.87 & 46.61
        \\
        $= 3$ & 0.1178 & 0.0913 & 0.1046 & 27.82 & 54.65 & 63.93 & 74.38
        \\
        $\leq 5$ & 0.0624 & 0.0553 & 0.0588 & 33.88 & 64.62 & 74.53 & 84.51
        \\
        $\leq 10$ & \textbf{0.0303} & \textbf{0.0331} & \textbf{0.0317} & \textbf{43.25} & \textbf{76.10} & \textbf{85.20} & \textbf{92.81}
        \\
        \noalign{\hrule height 0.85pt}
    \end{tabular}}
\end{table*}

\paragraphdot{Discussion}
As shown in Table~\ref{table:supp_analysis}, the geometric quality of sparse-view reconstructions degrades as the number of input images decreases. While reconstructions with $\leq 10$ images closely approach the dense-view pseudo ground truth, the gap substantially increases under fewer views, with single-view reconstructions exhibiting the largest deviation. This trend is consistent with the accuracy drop observed in the main paper, where limited geometric context contributes to alignment ambiguity.

\subsection{Evaluation by Scene View Type}
\label{sec:supp_analyses_interior_exterior_only_evaluation}

We analyze our method separately on interior and exterior scenes from the C3 dataset~\cite{C3Po}. As shown in Tables~\ref{table:camera_pose_image_floorplan_int_ext} and~\ref{table:correspondence_image_floorplan_int_ext}, our method achieves notably higher accuracy on interior scenes. We attribute this gap to inherent challenges of exterior scenes, where 3D reconstructions are noisier and often include far-away points from distant background regions. These factors introduce artifacts into the resulting density map, making cross-modal correspondence estimation more challenging.

\begin{table*}[!t]
    \centering
    \caption{\textbf{In-the-wild camera pose estimation by scene view type.} Our method is more accurate on interior scenes than on exterior scenes. The ``All'' row reports the aggregated accuracy.}
    \label{table:camera_pose_image_floorplan_int_ext}
    \vspace{-0.75mm}
    \renewcommand{\arraystretch}{1.1}
    \setlength{\tabcolsep}{5pt}
    \resizebox{\textwidth}{!}{
        \begin{tabular}{l cccc c ccc c c}
        
        \noalign{\hrule height 0.85pt}
        & \multicolumn{4}{c}{Angular Recall\,$\uparrow$} & \!\!\! & \multicolumn{3}{c}{Positional Recall\,$\uparrow$} & \!\!\! & \!\!\!Ang.\,\&\,Pos.\,$\uparrow$\!\!\!
        \\ 
        
        \cline{2-5}
        \cline{7-9}
        \cline{11-11}
        Type & \,@\,$5^\circ$ & \,@\,$10^\circ$ & \,@\,$20^\circ$ & \,@\,$30^\circ$ & \!\!\! & \,@\,$5\%$ & \,@\,$10\%$ & \,@\,$20\%$ & \!\!\! & \!\!\,@\,($30^\circ,20\%$)\!\!
        \\
        
        \noalign{\hrule height 0.7pt}
        Interior & \textbf{73.84} & \textbf{81.24} & \textbf{83.49} & \textbf{87.11} & \!\!\! & \textbf{64.84} & \textbf{81.22} & \textbf{92.62} & \!\!\! & \textbf{84.89}
        \\
        Exterior & 62.96 & 73.95 & 79.01 & 81.58 & \!\!\! & 45.66 & 59.75 & 69.91 & \!\!\! & 69.36
        \\
        \hline
        All & 65.91 & 75.93 & 80.23 & 83.08 & \!\!\! & 50.87 & 65.58 & 76.08 & \!\!\! & 73.58
        \\
        \noalign{\hrule height 0.85pt}
    \end{tabular}}
    \vspace{-1mm}
\end{table*}

\begin{table*}[!t]
    \centering
    \caption{\textbf{In-the-wild correspondence estimation by scene view type.} Our method is more accurate on interior scenes than on exterior scenes. The ``All'' row reports the aggregated accuracy.}
    \label{table:correspondence_image_floorplan_int_ext}
    \vspace{-0.75mm}
    \renewcommand{\arraystretch}{1.1}
    \setlength{\tabcolsep}{7pt}
    \resizebox{\textwidth}{!}{
        \begin{tabular}{l cccccc c}
        
        \noalign{\hrule height 0.85pt} 
        & \multicolumn{6}{c}{PCK\,$\uparrow$} & 
        \\ 
        
        \cline{2-7}
        Type & \,@\,1\%\, & \,@\,3\%\, & \,@\,5\%\, & \,@\,10\%\, & \,@\,15\%\, & \,@\,30\%\, & \,RMSE\,$\downarrow$\,
        \\
        
        \noalign{\hrule height 0.7pt}

        Interior & \textbf{24.66} & \textbf{63.47} & \textbf{74.43} & \textbf{84.18} & \textbf{87.06} & \textbf{95.09} & \textbf{0.0550} 
        \\
        Exterior & 18.84 & 56.14 & 68.17 & 78.36 & 82.37 & 88.73 & 0.0860
        \\
        \hline
        All & 20.42 & 58.13 & 69.87 & 79.94 & 83.64 & 90.45 & 0.0776 
        \\
        \noalign{\hrule height 0.85pt}
    \end{tabular}}
    \vspace{-2mm}
\end{table*}

\begin{table*}[!t]
    \centering

    \caption{\textbf{Amortized inference time on C3~\cite{C3Po}.}
    We report the amortized end-to-end processing time per image, decomposed into four stages. Ours ($\leq N$) processes input photos in chunks of at most $N$ images. Since the number of available photos varies per scene, the actual chunk size differs accordingly. For reference, we also report the mean chunk size.}
    \label{table:supp_inference_time}
    \vspace{-0.75mm}
    
    \renewcommand{\arraystretch}{1.1}
    \setlength{\tabcolsep}{4.25pt}
    \resizebox{\textwidth}{!}{
        \begin{tabular}{l c cccc c}
        
        \noalign{\hrule height 0.85pt} 
        & Mean & \multicolumn{4}{c}{Amortized Inference Time per Stage\,$\downarrow$\,} & 
        \\ 
        
        \cline{3-6}
        Method & Chunk Size & 3D Scene & Density Map & Correspondence & Camera Pose & \;Total\,$\downarrow$\,\;
        \\

        \hline
        C3Po~\cite{C3Po} & $1.0$ & --- & --- & 59.6\,ms & 135.8\,ms & 195.4\,ms 
        \\ 
        
        \noalign{\hrule height 0.7pt}
        Ours ($= 1$) & $1.0$ & 158.8\,ms & 20.2\,ms & 17.2\,ms & 0.2\,ms & 196.4\,ms 
        \\ 
        Ours ($\leq 10$) & $9.5$ & \textbf{121.8\,ms} & \textbf{20.0\,ms} & 1.9\,ms & \textbf{0.0\,ms} & \textbf{143.7\,ms}
        \\ 
        Ours ($\leq 150$) & $66.4$ & 200.2\,ms & 22.8\,ms & \textbf{0.4\,ms} & \textbf{0.0\,ms} & 223.4\,ms
        \\ 
        \noalign{\hrule height 0.85pt}
    \end{tabular}}
\end{table*}

\subsection{Inference Time Comparison}
\label{sec:supp_analyses_inference_time_comparison}
We compare the end-to-end inference time of our method against C3Po~\cite{C3Po} on a single NVIDIA RTX A6000 GPU, with results summarized in Table~\ref{table:supp_inference_time}. To reflect real-world deployment efficiency, we report the \textit{amortized cost} per image. For a detailed analysis, we decompose the inference time into four stages: (i) gravity-aligned 3D scene reconstruction, (ii) density map extraction, (iii) correspondence estimation, and (iv) camera pose estimation in the floorplan coordinate frame.

Our approach achieves a substantial speed advantage over C3Po in the latter two stages. During correspondence estimation, C3Po relies on a heavy ViT-L backbone for processing every single photograph. In contrast, we use a lightweight ViT-B backbone (initialized from DINOv3~\cite{DINOv3}) and estimate correspondences \textit{once} per density map, thereby effectively amortizing the cost across up to $N$ images. During camera pose estimation in the floorplan frame, C3Po solves epipolar geometry to produce candidate camera poses and selects the one closest to the ground truth. In contrast, our method estimates a 2D similarity transform and applies it to the reconstructed camera poses, incurring only negligible overhead.

We evaluate our method with maximum chunk sizes of $N \in \{ 1, 10, 150 \}$ and identify Ours ($\leq 10$) as the sweet spot, which is notably faster than C3Po (143.7\,ms vs. 195.4\,ms). At $N=1$, each input photo is processed sequentially and therefore the GPU is severely underutilized. At $N=150$, the quadratic complexity of transformer layers in the 3D reconstruction model becomes the dominant bottleneck. At $N=10$, these overheads are balanced, yielding the best trade-off.

%% file: sec/supp_6_limitation.tex
\setcounter{figure}{0}
\setcounter{table}{0}
\renewcommand\thefigure{G.\arabic{figure}}
\renewcommand\thetable{G.\arabic{table}}

\section{Limitations}
\label{sec:supp_limitations}

\begin{figure}[t!]
  \centering
  \includegraphics[width=\linewidth]{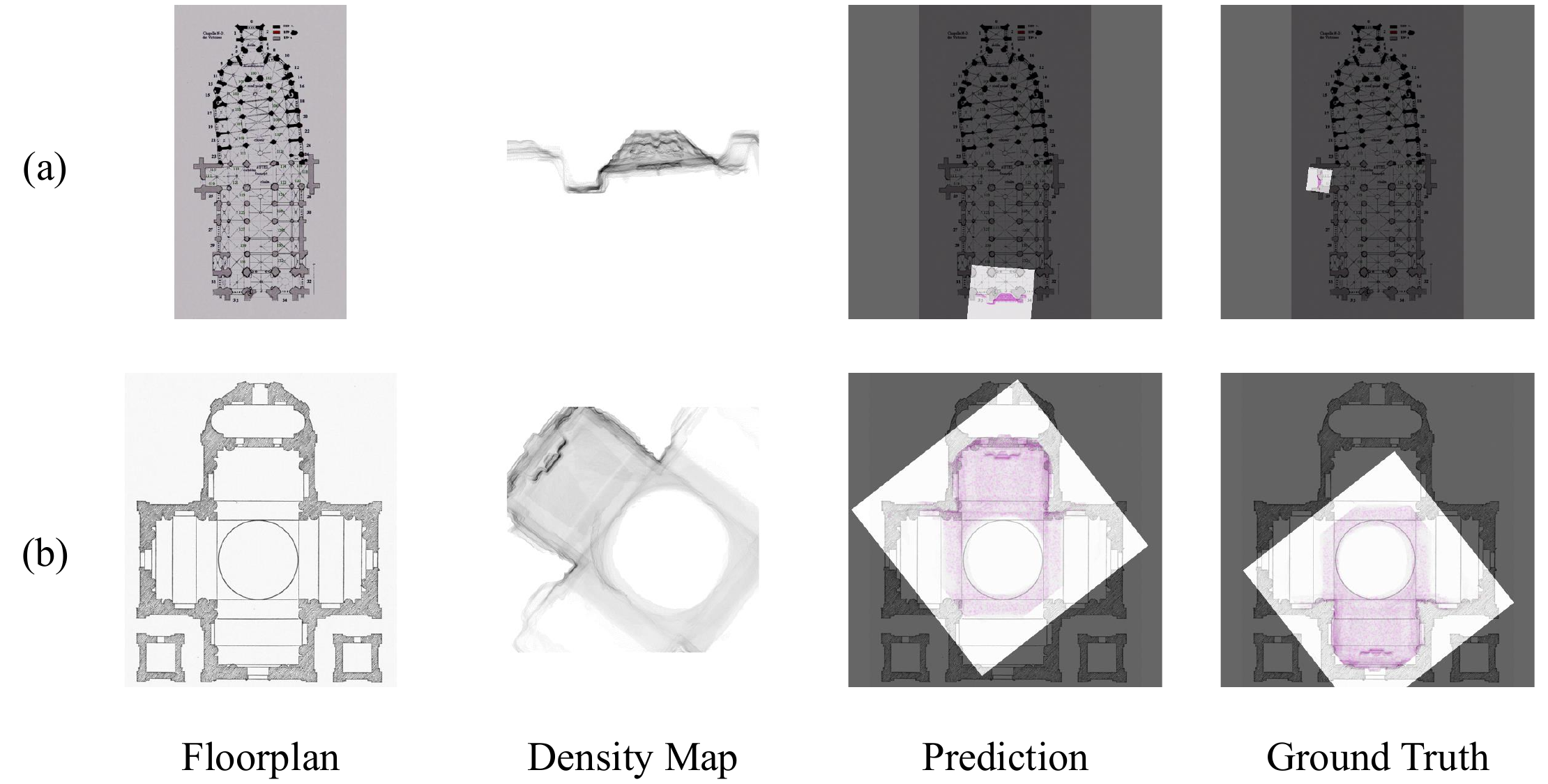}
  \vspace{-3mm}
  \caption{\textbf{Failure cases.} (a) Structural ambiguity. (b) Symmetrical ambiguity. For each row, we show the reference floorplan, extracted density map, predicted floorplan alignment, and ground-truth alignment. For visualization purposes, pink dots overlaid on the density map serve as reference points to visualize alignment quality (the same dots are illustrated in both alignment results).}
  \label{fig:supp_failure_cases}
  \vspace{-1mm}
\end{figure}

In Section~\ref{sec:conclusion} of the main paper, we noted that errors from 3D reconstruction models can degrade the quality of subsequent density maps and floorplan alignment. Beyond this, our method can also fail even with accurate 3D reconstructions, due to inherent ambiguities in cross-modal correspondence between density maps and floorplans.

Figure~\ref{fig:supp_failure_cases} illustrates two such failure modes. The first arises from \textit{structural ambiguity}, where the density map captures only a local region (\eg, a doorway in row (a)) that matches multiple visually similar structures on the floorplan. Without sufficient surrounding context, the model cannot uniquely determine which of these structures the density map corresponds to, which results in incorrect alignments. The second arises from \textit{symmetrical ambiguity}, where the density map captures a partially observed symmetric structure (\eg, three arms of a cross-shaped structure in row (b)). In such cases, multiple alignments can satisfy the local geometry equally well, and the model may choose an orientation that differs from the ground truth by $180^{\circ}$.

Both failure modes stem from a lack of global context in the density map. As such, they tend to be more pronounced under sparse-view settings (Section~\ref{sec:supp_analyses_sparse_view_3d_recon}), where the density map captures only a partial view of the scene. While our method already remains effective in most scenarios, addressing these ambiguities is an interesting direction for future work. 

%% file: sec/supp_7_application.tex
\setcounter{figure}{0}
\setcounter{table}{0}
\renewcommand\thefigure{H.\arabic{figure}}
\renewcommand\thetable{H.\arabic{table}}

\section{Applications}
\label{sec:supp_applications}

\begin{figure}[t!]
  \centering
  \includegraphics[width=\linewidth]{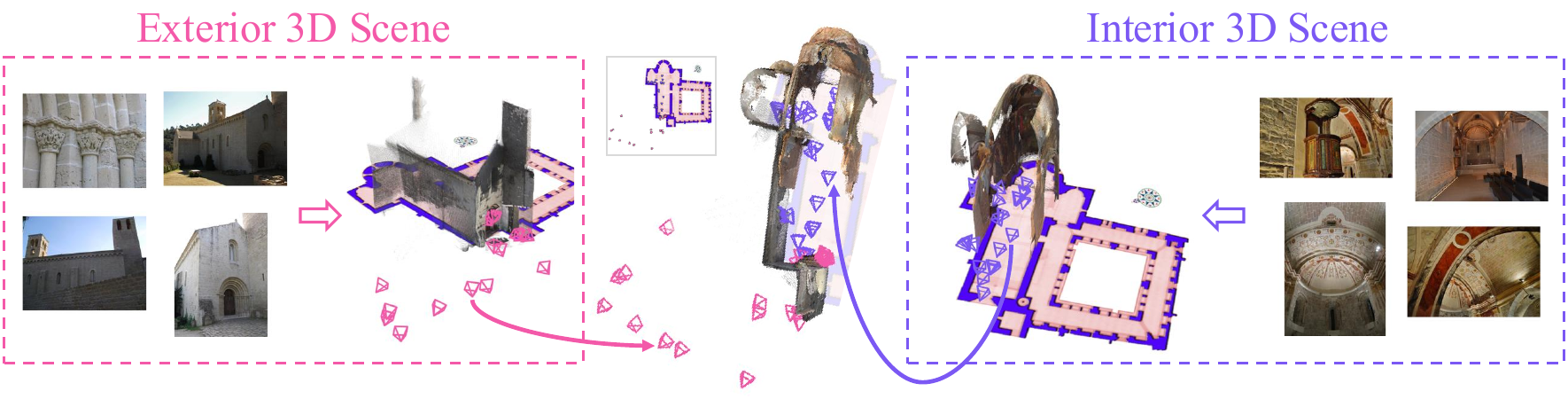}
  \vspace{-4.5mm}
  \caption{\textbf{Alignment of interior and exterior 3D scenes.} Using a reference floorplan as the shared geometric anchor, our method can align reconstructions from interior and exterior photos into a unified global coordinate system, despite minimal visual overlap and large viewpoint differences.}
  \label{fig:supp_interior_exterior_3d_alignment}
  \vspace{-1mm}
\end{figure}

\paragraphdot{Interior-Exterior 3D Scene Alignment}
State-of-the-art 3D foundation models~\cite{VGGT,Pi3} often fail to produce a coherent global reconstruction when jointly processing interior and exterior photos, primarily due to drastic viewpoint changes and the lack of sufficient visual overlap. By using a reference floorplan as a geometric anchor, our method can align separate reconstructions into a unified global coordinate system. We further showcase this application over reconstructed interior and exterior 3D scenes capturing the Monestir de Sant Benet de Bages (a medieval monastic complex in Catalonia) in Figure~\ref{fig:supp_interior_exterior_3d_alignment}.

\begin{figure}[t!]
  \centering
  \includegraphics[width=\linewidth]{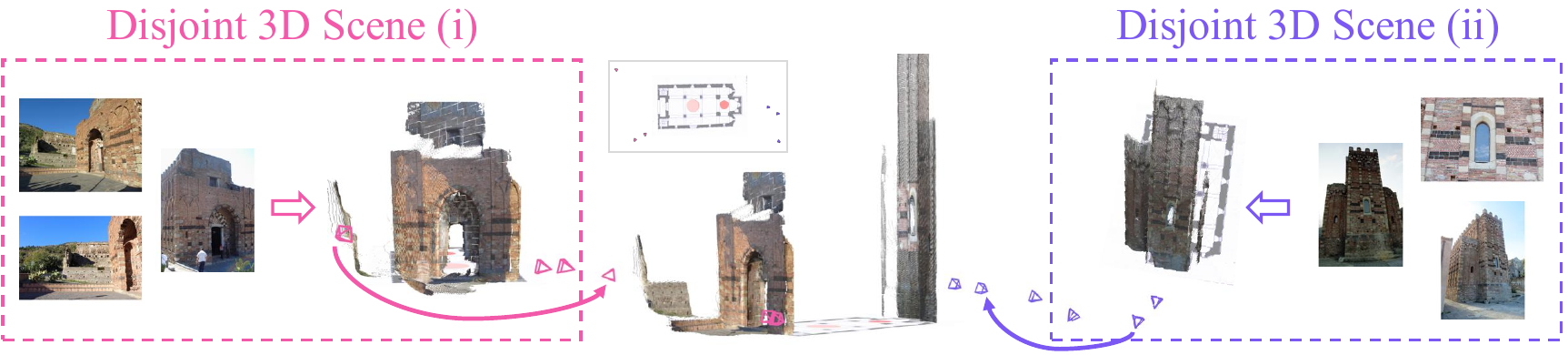}
  \vspace{-4.5mm}
  \caption{\textbf{Alignment of disjoint 3D scenes.} Leveraging a reference floorplan as the shared geometric bridge, our method can align reconstructions from disjoint image collections into a unified global coordinate system, even without visual overlapping.}
  \label{fig:supp_disjoint_3d_alignment}
  \vspace{-1mm}
\end{figure}

\paragraphdot{Disjoint 3D Scene Alignment}
Similarly, 3D foundation models frequently fail to register disjoint image collections that lack overlapping regions. Our method can address this by independently aligning disjoint reconstructions into a unified global coordinate system, as illustrated over partial reconstructions capturing Santi Pietro e Paolo d’Agrò (a church in Italy) in Figure~\ref{fig:supp_disjoint_3d_alignment}.

%% file: sec/supp_8_broader_impacts.tex
\setcounter{figure}{0}
\setcounter{table}{0}
\renewcommand\thefigure{I.\arabic{figure}}
\renewcommand\thetable{I.\arabic{table}}

\section{Broader Impacts}
\label{sec:supp_broader_impacts}

Our work enables localizing in-the-wild photographs within reference floorplans, which can support a wide range of applications, including scene navigation, augmented or virtual reality, and architectural understanding. By aligning unconstrained photo collections to floorplans, our method also facilitates downstream tasks such as interior-exterior 3D scene alignment and disjoint 3D scene alignment, which can be valuable for large-scale 3D reconstruction in challenging scenarios.

We do not expect significant negative societal impacts from this work. Our method operates on photographs and architectural floorplans, which are widely available and commonly used in geographic information systems. Nonetheless, as with any localization technology, responsible use is essential when applying it to private spaces. We encourage users to follow established guidelines when deploying our localization method in real-world scenarios.